\documentclass{article}

\PassOptionsToPackage{numbers,square,sort}{natbib} %%added by pippo
\usepackage[eandd,preprint]{neurips_2026}

\usepackage[utf8]{inputenc} % allow utf-8 input
\usepackage[T1]{fontenc}    % use 8-bit T1 fonts
\usepackage{hyperref}       % hyperlinks
\usepackage{url}            % simple URL typesetting
\usepackage{booktabs}       % professional-quality tables
\usepackage{amsfonts}       % blackboard math symbols
\usepackage{nicefrac}       % compact symbols for 1/2, etc.
\usepackage{microtype}      % microtypography
\usepackage{xcolor}         % colors
\usepackage{graphicx}
\usepackage{array} % in preamble
\usepackage{placeins}

\usepackage[table]{xcolor}
\usepackage{float}
\usepackage{afterpage}

\usepackage[most]{tcolorbox}
\usepackage{enumitem}
\usepackage{xcolor}
\usepackage[table]{xcolor}
\usepackage{graphicx}
\usepackage{subcaption}
\usepackage{paralist}
\usepackage{amssymb}

\tcbset{
    expbox/.style={
        enhanced,
        breakable,
        colback=gray!3,
        colframe=black!25,
        boxrule=0.5pt,
        arc=2pt,
        left=6pt,
        right=6pt,
        top=6pt,
        bottom=6pt,
        before skip=7pt,
        after skip=7pt,
        fonttitle=\bfseries,
        coltitle=black,
        colbacktitle=gray!12,
        attach boxed title to top left={xshift=5pt,yshift=-2pt},
        boxed title style={
            colback=gray!12,
            colframe=black!20,
            boxrule=0.4pt,
            arc=2pt,
            left=5pt,
            right=5pt,
            top=2pt,
            bottom=2pt
        }
    }
}

\definecolor{boxbg}{RGB}{248,250,252}       % very light slate
\definecolor{boxframe}{RGB}{148,163,184}   % muted slate border
\definecolor{boxtitlebg}{RGB}{226,232,240} % soft title background
\definecolor{boxtitle}{RGB}{30,41,59}      % dark slate text

\tcbset{
    metricbox/.style={
        enhanced,
        breakable,
        colback=boxbg,
        colframe=boxframe,
        boxrule=0.55pt,
        arc=2pt,
        left=6pt,
        right=6pt,
        top=6pt,
        bottom=6pt,
        before skip=6pt,
        after skip=6pt,
        fonttitle=\bfseries,
        coltitle=boxtitle,
        colbacktitle=boxtitlebg,
        attach boxed title to top left={xshift=5pt,yshift=-2pt},
        boxed title style={
            colback=boxtitlebg,
            colframe=boxframe,
            boxrule=0.45pt,
            arc=2pt,
            left=5pt,
            right=5pt,
            top=2pt,
            bottom=2pt
        }
    }
}

\title{
\texorpdfstring{
    \raisebox{-1.9ex}{\includegraphics[height=1.6cm]{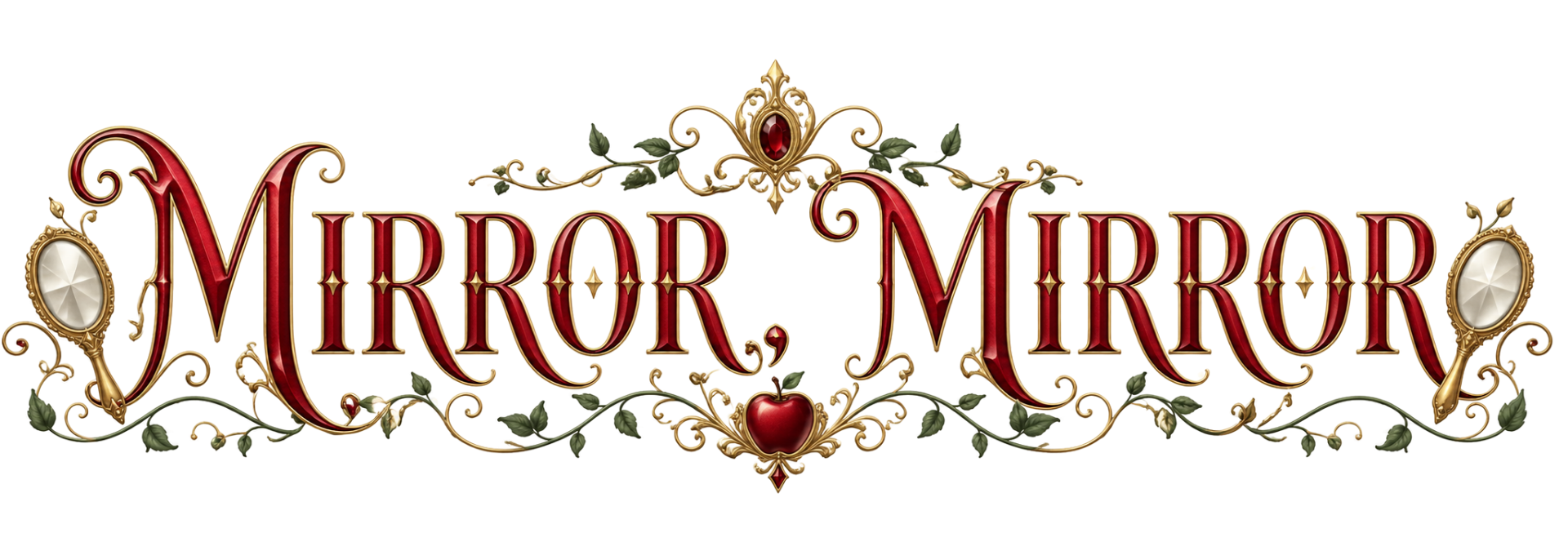}}
    \hspace{-0.3em}
    on the Wall: Can VLM Agents Tell\\[-0.1em]
    Who They Are at All?
}{
    Mirror, Mirror on the Wall: Can VLM Agents Tell Who They Are at All?
}
}
\author{%
  Filippo Ziliotto\thanks{Equal contribution.} \\
  University of Padova \\
  Fondazione Bruno Kessler \\
  \And
  Ciro Beneduce\footnotemark[1] \\
  University of Trento \\
  Fondazione Bruno Kessler \\
  \And
  Bruno Lepri \\
  Fondazione Bruno Kessler \\
  \AND
  Luciano Serafini \\
  Fondazione Bruno Kessler \\
  \And
  Massimiliano Luca\thanks{Equal supervision.} \\
  Fondazione Bruno Kessler \\
  \And
  Tommaso Campari\footnotemark[2] \\
  Fondazione Bruno Kessler \\
}

\begin{document}

\maketitle

\begin{abstract}
In the animal kingdom, mirror self-recognition is a canonical probe of higher-order cognition, emerging only in some species. We ask whether an analogous functional capability emerges in embodied vision-language model (VLM) agents: \textit{can they recognize themselves in a mirror?}
We introduce a controlled 3D benchmark where a first-person VLM agent must infer a hidden body attribute from its reflection and select the matching target, while avoiding self–other misattribution. To separate mirror-grounded self-identification from shortcuts, we test mirror removal, misleading cues, and occluded reflections. We also evaluate the decision process through mirror seeking, temporal ordering, self-attribution, and reasoning–action consistency.
Our experiments show that mirror-based self-identification emerges mainly in stronger VLMs. These models can use reflected evidence for action, whereas weaker models often inspect the mirror but fail to extract self-relevant information or misattribute their reflection. Language–vision conflict further shows that self-referential language alone is not evidence of grounded self-identification.
Overall, mirror-based evaluation provides a diagnostic for whether embodied self-grounding is causally rooted in perception and action rather than priors, prompt compliance, or confabulation.


\end{abstract}

\section{Introduction}

\begin{figure}[t]
    \centering
    \includegraphics[width=0.7\linewidth]{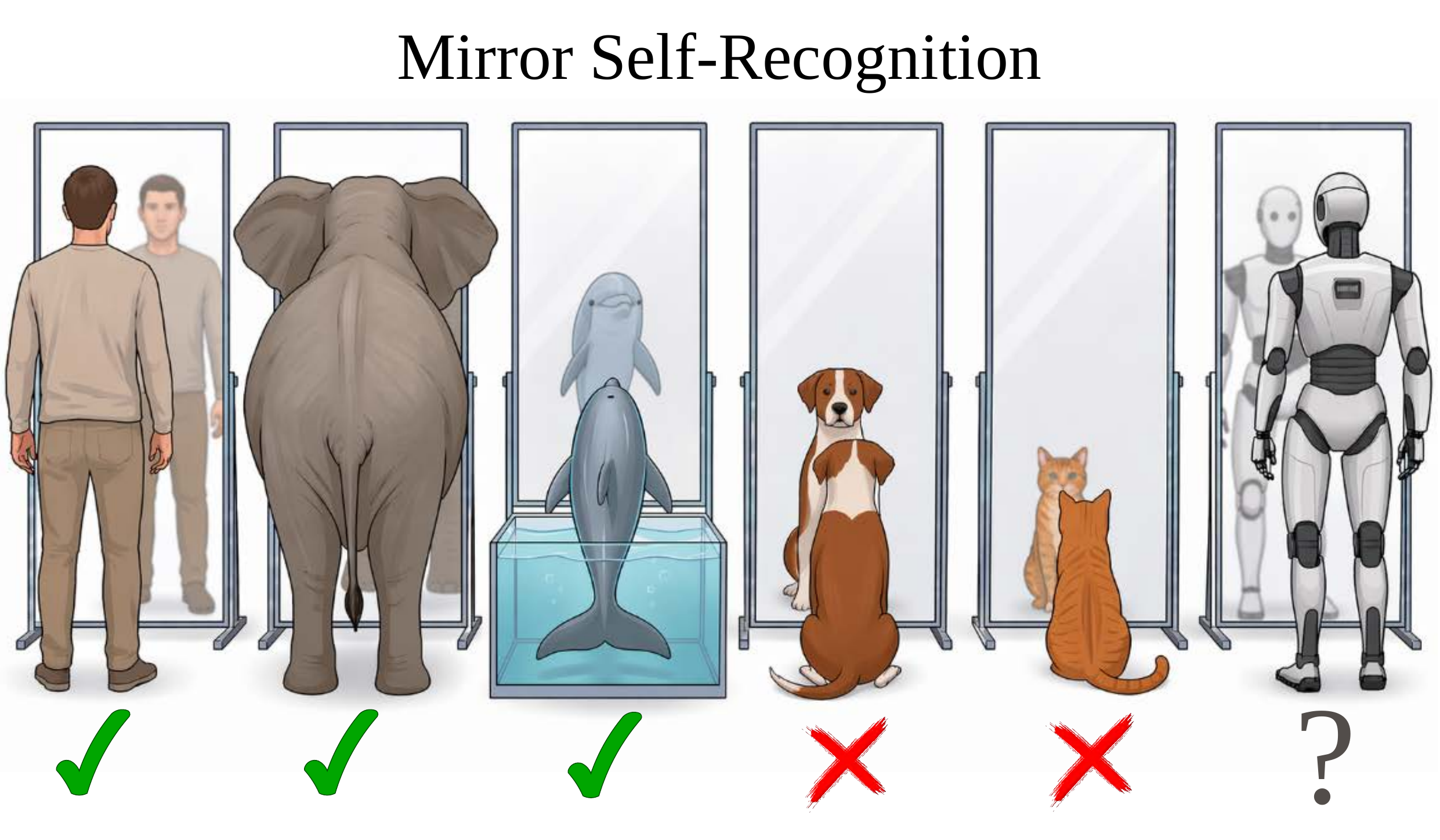}
    \caption{\textbf{Mirror self-recognition as a test of embodied self-grounding.}
    Across species, mirror self-recognition emerges selectively: humans, elephants, and dolphins can use mirror reflections to guide self-directed behavior, while other animals fail or rely on simpler strategies. We ask whether a similar capability emerges in vision-language model (VLM) agents. In our setting, an embodied agent must infer a hidden body attribute from a mirror and act accordingly. Success requires not only perceiving the reflection, but also attributing it to the self rather than to another entity.}
    \label{fig:teaser}
    \vspace{-0.5cm}
\end{figure}

Vision-language models (VLMs) are increasingly used as decision-making modules for embodied agents that must perceive, reason, and act under partial observability. This makes visual grounding essential: an agent must not only describe what it sees, but use visual evidence to choose actions. Despite strong progress in recognition, instruction following, navigation, and multimodal reasoning, current VLMs still suffer from hallucination, weak visual grounding, and failures in spatial reasoning \citep{Zhang2023Vision-Language,hu2019you,Rahmanzadehgervi2024Vision,Chen2024Are}. These limitations become especially important when the object of reasoning is not an external object, but the agent itself.

Self-identification is a basic requirement for embodied agency. An agent must distinguish its own body, state, and actions from those of other agents and objects. Under egocentric perception, however, self-relevant information is often outside the camera view and cannot be accessed directly. Mirrors provide a controlled way to test this ability: they reveal hidden body information while requiring the agent to attribute the reflected entity to itself and use that attribution for action.

Mirror self-recognition has long been studied in comparative cognition. Classic work showed that chimpanzees can use mirror reflections to guide self-directed behavior \citep{gallup1970chimpanzees}, and later studies reported related behavior in dolphins, elephants, and other species \citep{plotnik2006selfrecognition,morrison2018precocious}. Yet the interpretation of mirror-test success remains debated. It may reflect self-awareness, but it may also arise from sensorimotor matching, learned visual-proprioceptive associations, or task-specific strategies \citep{Brandl2018The,zhou2015monkeys}. This caution should also apply to embodied agents: fluent self-referential language or correct final actions are not sufficient evidence of grounded self-identification.

In this paper, we ask whether embodied VLM agents can perform mirror-guided self-identification (Fig.~\ref{fig:teaser}). We define this as a functional capability: an agent must use reflected visual evidence to infer a hidden attribute of its own body and then use that information to guide goal-directed behavior. This framing tests whether self-referential behavior is grounded in perception and action rather than produced by priors, prompt compliance, or confabulated explanations.

To study this question, we introduce a controlled benchmark\footnote{Code \& Benchmark are available at https://anonymous.4open.science/r/self-awareness-D446} in which a VLM-controlled agent observes the environment from a first-person perspective and must select the target object matching a hidden body attribute, i.e. its own body color. The attribute is not directly visible and can only be inferred through a mirror. We further introduce diagnostic interventions that remove the mirror, provide misleading linguistic cues, or degrade the reflection through occlusion. These conditions allow us to distinguish mirror-grounded self-identification from shortcut behavior, language bias, scene priors, and guessing.

A key aspect of our benchmark is that it evaluates both outcomes and decision processes. Target-selection accuracy indicates whether the agent acts correctly, but not whether the decision was grounded in mirror evidence. We therefore also measure mirror consultation, mirror gaze duration, temporal ordering between mirror observation and action, self-attribution in language, and confabulation. This process-level evaluation is necessary because embodied VLMs can appear competent while relying on weak grounding or non-visual shortcuts \citep{hu2019you,Rahmanzadehgervi2024Vision,Chen2024Are}.

Our experiments show that mirror-guided self-identification can emerge in stronger VLM agents but remains fragile. Some models use reflected evidence to guide action, whereas others inspect the mirror without extracting the relevant self-attribute. We also find that self-referential language is not a reliable proxy for grounded self-identification: models often produce coherent first-person explanations even when mirror evidence is absent, misleading, or degraded. These results suggest that mirror-based evaluation is a useful diagnostic for embodied self-grounding in VLM agents.

Our contributions are as follows:
\begin{inparaenum}[(i)]
    \item we frame mirror-guided self-identification as a functional embodied capability for VLM agents and introduce a controlled 3D benchmark requiring agents to infer hidden self-relevant attributes from mirror evidence and act accordingly;
    \item we design interventions that remove mirror evidence, introduce misleading language, and degrade reflections to test whether behavior depends on visual self-grounding;
    \item we propose process-level metrics covering task success, evidence seeking, temporal ordering, self-attribution, and confabulation;
    \item we show that mirror-guided self-identification emerges mainly in stronger VLMs, whereas weaker agents often consult the mirror but fail to reliably use reflected evidence, frequently mistaking themselves for another agent.
\end{inparaenum}
\vspace{-0.3cm}
\section{Related Work}
\vspace{-0.3cm}

\textbf{Vision-Language Models for Embodied Reasoning.}
Vision-language models (VLMs) connect visual perception with language-conditioned reasoning and have shown strong zero-shot and few-shot performance across recognition, captioning, visual question answering, navigation, and multimodal reasoning \citep{Zhang2023Vision-Language}. 
In embodied settings, they can map visual observations and instructions into high-level plans or actions, but strong multimodal performance does not necessarily imply grounded visual reasoning: VLMs still suffer from hallucination, weak modality alignment, textual priors, and failures on spatially simple or perceptually grounded tasks \citep{Li2025Self-Rewarding, Zhou2024Calibrated, Wang2024Enhancing, xu2024lvlm, Liu2023MMBench:, Rahmanzadehgervi2024Vision}. 
Recent work improves grounding through perception-language decomposition, self-rewarding, calibration, navigation history, and action-conditioned reasoning \citep{Li2025Self-Rewarding, Zhou2024Calibrated, Wang2024Enhancing, hu2019you, yang2019embodied}. 
Our work shifts this focus from grounding external objects to grounding self-relevant visual information.

\textbf{Multimodal Benchmarks, Shortcuts, and Process-Level Evaluation.}
Recent multimodal benchmarks evaluate object hallucination, referential comprehension, compositional reasoning, self-consistency, and visual-language competence \citep{xu2024lvlm, Liu2023MMBench:, Yue2024SC-, wang2025enhancing, Zhou2024Calibrated, Soderlund2022When}. 
These benchmarks show that high accuracy can still arise from shortcuts, including language priors, dataset regularities, or memorized correlations rather than genuine image use \citep{Chen2024Are}. 
Vision-centric benchmarks such as NaturalBench and MMStar therefore design tasks that require image-grounded reasoning \citep{Li2024NaturalBench:, Chen2024Are}. 
This issue is central to embodied self-identification: an agent may choose the correct target without using the mirror, produce self-referential language without perceptual evidence, or misattribute its reflection to another entity. 
%We therefore evaluate not only final task success, but also evidence seeking, mirror consultation, temporal ordering, self-attribution, and confabulation.

\textbf{Mirror Self-Recognition in Animals and Humans.}
Mirror self-recognition has long been used as a behavioral probe of self-related processing in comparative cognition. 
The classic mark-test paradigm reported that chimpanzees can use mirrors to guide self-directed behavior \citep{gallup1970chimpanzees}, and later work reported related findings in chimpanzees, orangutans, dolphins, elephants, magpies, and other species, while also documenting important species-level differences \citep{suarez1981chimpanzees,gallup1982selfawareness,marten1994dolphin,plotnik2006selfrecognition,prior2008magpie, zhou2015monkeys, morrison2018precocious}. 
The interpretation of these results remains contested: mirror-test success may indicate self-awareness or self-concept, but it may also reflect sensorimotor matching, kinesthetic-visual correspondence, learned associations, or task-specific strategies \citep{mitchell1997kinesthetic,morin1989gallupsmirrors,gallup2002mirrortest,bekoff2004reflections,chang2017monkeymsr}. 
We adopt this cautious interpretation and test mirror-guided self-identification as a functional capacity rather than as evidence of philosophical self-awareness.

\textbf{Artificial Self-Recognition and Embodied Agency.}
Artificial self-recognition has been studied in robotics and cognitive architectures through kinesthetic-visual matching, body-model learning, inner speech, multisensory integration, and sensorimotor contingencies \citep{Hoffmann2020Robot, Pipitone2021Robot, Zeng2017Toward}. 
These approaches relate to psychological and developmental accounts that distinguish first-order sensorimotor matching from more reflective self-awareness, and that emphasize action--perception coupling in distinguishing self from other \citep{Mitchell1993Mental, Brandl2018The, steels2008robot, Shimada2022Multisensory}. 
Unlike systems explicitly engineered for self-recognition, we test whether a general-purpose VLM agent can infer a hidden body attribute from first-person mirror evidence and use it for goal-directed action.

\textbf{Positioning of This Work.}
Prior work shows that VLMs can perform well while relying on weak grounding, shortcuts, or textual priors, and that mirror self-recognition remains a contested but influential probe of self-related processing. 
We connect these threads by studying mirror-guided self-identification as a functional embodied capability. 
%Our setting requires agents to infer hidden self-relevant body attributes from reflected visual evidence and act on them. 
By measuring mirror consultation, temporal ordering, self-attribution, and confabulation, we distinguish mirror-grounded self-identification from shortcut behavior, prompt compliance, and misattribution of the agent's own reflection.
\section{Setup}
\label{sec:method}

\subsection{Motivation and Research Question} \label{subsec:Motivation_and _research_questions}

We study whether an embodied VLM agent can treat visual observations as self-referential evidence. Rather than addressing self-awareness in a philosophical sense, we focus on a functional question: can an agent infer a hidden property of its own body from a reflection and use it to guide action?

The task is designed so that self-relevant information is never directly visible from the egocentric view. The agent must instead recover it indirectly through a reflective surface, under conditions that introduce ambiguity, distractors, misleading language, or degraded visual evidence. This allows us to distinguish mirror-grounded self-identification from guessing, prompt following, scene priors, or post-hoc confabulation.

\subsection{Embodied Setting and Task} \label{subsec: Setting_and_task}

% The agent operates in a simulated 3D environment from a first-person perspective. At each step, it receives the current RGB observation, a compact action-position history, and minimal navigation context, and outputs one of four discrete actions: \textit{move\_forward}, \textit{turn\_left}, \textit{turn\_right}, or \textit{stop}.

% The prompt does not explicitly mention a mirror, and the reflective surface is not visually labeled. This design reduces semantic shortcuts based on the word ``mirror'' and encourages the agent to rely on perceptual and temporal evidence. Successful behavior requires the agent to infer that the reflected body corresponds to itself, for example through temporal consistency, action-contingent motion, and the causal relation between its own movements and the observed reflection.

% Across all conditions, the agent's body color is hidden from direct view and can only be inferred indirectly. The core task is to identify this hidden self-attribute and, when required, navigate to the colored cube that matches it. Success therefore requires not only recognizing a color, but correctly attributing the reflected body to the self and using this attribution for action.

The agent operates in a simulated 3D environment from a first-person perspective. At each timestep \(t \in \{1,\ldots,T\}\), it receives \(x_t = \bigl(I_t, h_{<t}, p_t, q\bigr)\) and outputs \(y_t = \bigl(a_t, \hat{c}^{\mathrm{self}}_t, z_t\bigr)\), where \(I_t\) is the current RGB observation, \(h_{<t}\) is a compact action-position history, \(p_t\) is the navigation state, \(q\) is the task instruction, and $a_t \in \{\texttt{move\_forward}, \texttt{move\_backward}, \texttt{turn\_left}, \texttt{turn\_right}, \texttt{done}\}$ is the chosen action. The variable \(\hat{c}^{\mathrm{self}}_t\) denotes the current self-attribution, while \(z_t\) denotes the task-dependent decision field: cube selection in \(E_1\)–\(E_4\) and final self-attribution in \(E_5\).

Across all conditions, the agent's true body color \(c^\star\) is never directly visible from the egocentric view and must be inferred indirectly from perceptual evidence in the scene. The reflective mirror surface is not visually labeled, so successful behavior requires grounding self-identification in action-contingent and temporally consistent visual evidence, rather than relying only on priors or scene regularities.
A visual example of all the experimental setups is shown in Figure~\ref{fig:experiment_setup}, while details on the environment can be found in Appendix \ref{sec:appendix-protocol}.

\subsection{Experimental Conditions} \label{Experimental Conditions}

\textbf{E1: Active Mirror-Based Self-Identification.}
The agent is placed in a room with an unlabeled reflective surface and multiple colored cubes. Its body color is hidden from direct view. To solve the task, the agent must explore, discover the reflective evidence, infer its own body color, and navigate to the matching cube. This tests whether an embodied agent can identify itself through reflection rather than through chance or scene priors.

\textbf{E2: Self-Identification Without Perceptual Evidence.}
This condition matches E1, but the reflective surface is removed. The agent must still select the cube corresponding to its own body color, although no visual evidence about this attribute is available. Performance should therefore collapse to chance; residual success indicates reliance on priors, prompt bias, scene regularities, or guessing.

\textbf{E3: Self-Identification Under Conflicting Language.}
This condition matches E1, but the agent receives a misleading textual statement about its own body color that conflicts with the mirror evidence. The task tests whether the agent privileges visual self-evidence over false linguistic information, separating perceptual grounding from prompt compliance.

\textbf{E4: Self--Other Disambiguation via Action.}
The agent is placed in a room with multiple other agents, ranging from one to six, which may have similar or identical body colors. All agents can appear both directly and through the reflective surface. The agent must identify which reflection is causally linked to its own actions and avoid misattributing another agent or reflection as itself, and then navigate to the cube matching its own body color.

\textbf{E5: Robust Open-Ended Self-Attribution Under Clutter.}
Unlike $E_1$--$E_4$, E5 is an open-ended exploration probe rather than a
cube-selection task. The agent explores a cluttered mirrored room with
distractors and must decide when to terminate with a final self-attribution.
This tests whether self-identification remains possible when mirror evidence is
partial, noisy, or intermittently occluded.

Formally, for each condition $e \in \{E_1,\dots,E_5\}$, let $\tau$ denote the first timestep at which the agent emits \texttt{done}, with the convention $\tau = T$ if no \texttt{done} action is produced.
Let $c^\star$ denote the true body color of the ego agent.
We define the final task decision as
\[
\scriptstyle
z_\tau^{(e)} =
\begin{cases}
\hat{c}^{\mathrm{cube}}_\tau, & e \in \{E_1,E_2,E_3,E_4\},\\[4pt]
\hat{c}^{\mathrm{self}}_\tau, & e = E_5,
\end{cases}
\]
where $\hat{c}^{\mathrm{cube}}_\tau$ is the final selected cube color and
$\hat{c}^{\mathrm{self}}_\tau$ is the final self-color attribution.

\begin{figure}[t]
    \centering
    \includegraphics[width=1.0\linewidth]{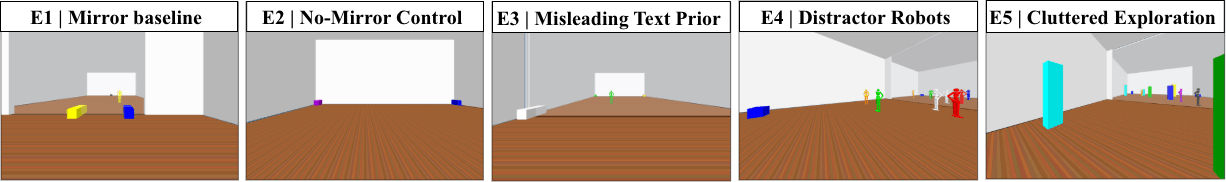}
    \caption{\textbf{Experimental settings.} Visual examples of the proposed conditions and 3D environments. At each timestep, the agent observes the current first-person view RGB frame, a compact textual history, and minimal navigation context. It then outputs a discrete navigation action, an image description, and an updated episode summary that is carried forward as context.}
    \label{fig:experiment_setup}
\end{figure}

\subsection{Design Rationale}

The task suite progressively probes embodied self-identification under increasing ambiguity. E1 tests the core capability, E2 removes perceptual evidence, E3 creates language--vision conflict, E4 introduces self--other ambiguity, and E5 tests robustness to degraded reflections. Together, these conditions ask whether VLM agents can recognize themselves in a mirror as a grounded functional capability, rather than merely produce correct answers or self-referential language.

\subsection{Behavioral and Process-Level Measures} \label{subsec:Metrics}

We evaluate both task outcomes and decision processes. Outcome metrics capture whether the agent selects the correct target, but this alone does not show whether the decision was grounded in mirror evidence. We therefore also measure whether the agent consults the mirror, whether mirror observation precedes action, whether the reflected body is attributed to the self, and whether the agent confabulates self-knowledge without evidence.
Generally, for episode $i$, let
\[
\tau_i=\min\{t:a_{i,t}=\texttt{done}\},
\]
with $\tau_i=T_i$ if no \texttt{done} action is produced. Let
$z^{(e)}_{i,\tau_i}$ denote the final task decision defined in Section \ref{Experimental Conditions}, let
$c_i^\star$ be the true ego-agent color, and let $m_{i,t}\in\{0,1\}$ indicate
whether mirror evidence is visible at timestep $t$.

%\paragraph{Task success metrics.}
\textbf{Task Success Accuracy (TSA)} measures whether the agent reaches the correct final task decision Formally:  \[
\scriptstyle
\mathrm{TSA}_i =
\begin{cases}
1, & \text{if } z^{(e)}_{i,\tau_i} = c_i^\star,\\
0, & \text{otherwise.}
\end{cases}
\]

\textbf{Time-to-Decision (TTD)} measures the number of timesteps before
termination:
\[
\scriptstyle
\mathrm{TTD}_i = \tau_i.
\] 
It helps separating immediate biased choices from decisions preceded by exploration.

%\paragraph{Mirror interaction metrics.}
%\textbf{MGD} measures the number of steps spent oriented toward the mirror. Low values indicate limited mirror use, while very high values may indicate uncertainty or failure to extract actionable evidence. 
\textbf{Mirror Consultation Rate (MCR)} measures whether the agent observes mirror evidence before committing to the final decision, distinguishing mirror-based decisions from visually unsupported ones. 
\[
\scriptstyle
\mathrm{MCR}_i =
\begin{cases}
1, & \text{if there exists } t < \tau_i \text{ such that } m_{i,t}=1,\\
0, & \text{otherwise.}
\end{cases}
\]
\textbf{Mirror-Then-Action Temporal Ordering (MTATO)} measures the fraction of successful episodes in which mirror consultation occurs before the final decision, testing whether correct behavior is temporally preceded by relevant evidence.
For successful episodes only, we define
\[
\scriptstyle
\mathrm{MTATO}_i =
\begin{cases}
1, & \text{if } \mathrm{TSA}_i = 1 \text{ and there exists } t < \tau_i \text{ such that } m_{i,t}=1,\\
0, & \text{if } \mathrm{TSA}_i = 1 \text{ and no such } t \text{ exists.}
\end{cases}
\]
We then report the mean of $\mathrm{MTATO}_i$ over successful episodes only.
%\textbf{SRL} measures whether the agent attends to its own reflected body rather than to other reflected objects or distractor agents.

%\paragraph{Language-based metrics.}
%\textbf{SRR} measures the frequency of first-person or self-referential expressions. 
\textbf{Correct Attribution At Least Once (CAAL)} measures whether the agent
ever states the correct self-color before termination:
\[
\scriptstyle
\mathrm{CAAL}_i =
\begin{cases}
1, & \text{if there exists } t < \tau_i \text{ such that } \hat{c}^{\mathrm{self}}_{i,t} = c_i^\star,\\
0, & \text{otherwise.}
\end{cases}
\]
%\textbf{EUL} measures uncertainty markers such as ``I think'' or ``it appears'', especially when evidence is incomplete or ambiguous. 
%\textbf{RCS} measures whether the transcript maintains a coherent mapping between the agent, its reflection, and the inferred body attribute. 
\textbf{Confabulation Rate (CR)} measures whether the agent commits to a
self-attribution before observing mirror evidence. Let
$t_i^{\mathrm{claim}}$ denote the first timestep before $\tau_i$ at which the
agent outputs a self-color different from \texttt{unknown}, and let
$t_i^{\mathrm{mirror}}$ denote the first timestep before $\tau_i$ at which
mirror evidence becomes visible, whenever these timesteps exist. We define
\[
\scriptstyle
\mathrm{CR}_i =
\begin{cases}
1, & \text{if } t_i^{\mathrm{claim}} \text{ exists and either } t_i^{\mathrm{mirror}} \text{ does not exist}\\
   & \text{or } t_i^{\mathrm{claim}} < t_i^{\mathrm{mirror}},\\
0, & \text{otherwise.}
\end{cases}
\]

Together, these metrics separate mirror-grounded self-identification from superficially correct behavior, guessing, prompt compliance, and post-hoc explanation.

%\begin{figure}[t]
   % \centering
   % \includegraphics[width=0.9\linewidth]{images/Fig_2_self_awereness.pdf}
   % \caption{\textbf{Experimental settings.} Visual examples of the proposed %conditions and 3D environments. At each timestep, the agent observes the %current first-person view RGB frame, a compact textual history, and %minimal navigation context. It then outputs a discrete navigation action, %an image description, and an updated episode summary that is carried %forward as context.}
%    \label{fig:experiment_setup}
%\end{figure}

\section{Results}
\label{sec:results}

We analyze the results with respect to the diagnostic questions and hypotheses defined in Sec.~\ref{Experimental Conditions}, as shown in Table~\ref{tab:all-results}. The central goal is not only to determine whether an agent reaches the correct final task decision but also to ensure that correct behavior is supported by mirror-grounded self-identification. We therefore interpret the results through the three metric groups introduced in Sec.~\ref{subsec:Metrics}: task-success metrics, mirror-interaction quality metrics, and text-based metrics. This distinction is crucial: an agent may succeed by chance or shortcut, may inspect the mirror without extracting the relevant self-attribute, or may produce fluent self-referential language without perceptual grounding. For additional technical details, extended metrics, and qualitative examples, refer to Appendix~\ref{sec:appendix}.

We selected eight different VLMs to span the current frontier of multimodal agents across major providers (Anthropic~\cite{anthropic2026claude46}, Google~\cite{Comanici2025Gemini2P}, OpenAI~\cite{singh2026openaigpt5card}, Mistral~\cite{liu2026ministral}, Qwen~\cite{qwen3technicalreport2025}) and a wide range of capability tiers, from compact open-weight models to flagship proprietary systems. This spread allows us to test whether mirror-guided self-identification scales with model strength or emerges only beyond a capability threshold. All VLM inference was performed via provider APIs; the only local compute requirement was a single workstation (128 GB RAM, NVIDIA RTX 2040 8 GB) used to run the MuJoCo~\cite{todorov2012mujoco} simulation and orchestrate API calls.

% We use TSA-C to denote complete task success, TSA for target-selection accuracy, MCR for mirror consultation rate, TTD for time-to-decision, MGD for mirror gaze duration, MTATO for mirror-then-act temporal ordering, SRR for self-reference rate, CAAL for color attribution accuracy in language, EUL for epistemic uncertainty language, RCS for referential coherence score, and CONFAB. for confabulation rate.

\begin{figure}[t]
    \centering
    \includegraphics[width=1\linewidth]{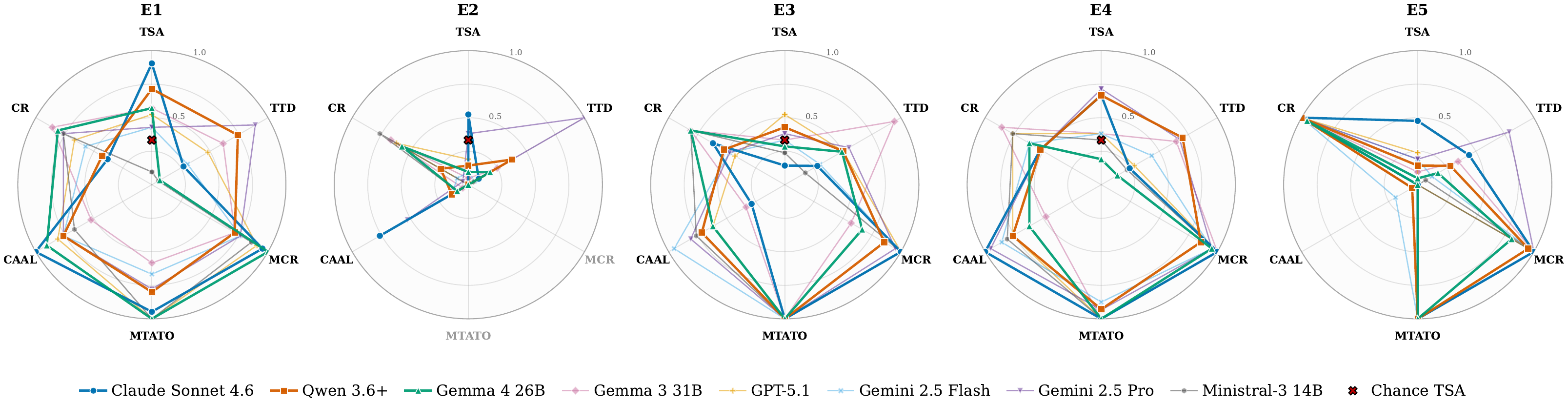}
    \caption{\textbf{Behavioral self-recognition profiles across task variants.}
Model comparison across experiments using TSA, TTD, MCR, MTATO, CAAL, and CR. Higher is better on all axes; therefore, TTD and CR are reversed for visualization. The red marker on the TSA axis indicates chance level.}
    \label{fig:spider_plot}
    \vspace{-0.3cm}
\end{figure}

\textbf{E1: Can an Embodied Agent Discover Who It Is by Looking at Its Reflection?}
E1 evaluates the core capability targeted by the benchmark: whether an embodied VLM agent can discover a reflective surface, infer its hidden body color from the reflected body, attribute that reflection to itself, and use the inferred self-attribute to guide navigation. The results in Table~\ref{tab:all-results} and Figure~\ref{fig:spider_plot} provide evidence that this capability can emerge, but only for some models and not as a uniform property of VLM agents.

From the perspective of task success, \texttt{Claude Sonnet 4.6} provides the strongest evidence of mirror-based self-identification, achieving TSA $=0.905 \pm 0.066$. \texttt{Qwen 3.6 Plus} also performs strongly, with TSA $=0.714 \pm 0.101$. These results show that some agents can solve the basic mirror-guided self-identification task. However, the capability is highly model-dependent: \texttt{Ministral 3 14B} reaches only TSA $=0.095 \pm 0.066$, indicating that the task remains difficult even when the mirror is available.

The mirror-interaction metrics reveal the main diagnostic dissociation. High mirror consultation does not guarantee grounded self-identification. \texttt{Gemma4 26B} consults the mirror in every episode (MCR $=1.000 \pm 0.000$) and has perfect temporal ordering (MTATO $=1.000 \pm 0.000$), but reaches only TSA $=0.571 \pm 0.111$. The same pattern is more pronounced for \texttt{Ministral 3 14B}, which has high MCR and perfect MTATO but very low TSA. Thus, the relevant failure is not merely failing to find the mirror, but failing to extract and use the reflected self-attribute.

% Text-based metrics further caution against treating self-referential language as evidence of self-grounding. SRR is high for most models, indicating frequent first-person or self-referential language. Yet CAAL and CONFAB. vary substantially. \texttt{Claude Sonnet 4.6} achieves perfect CAAL but also a high CONFAB. rate of $0.619 \pm 0.109$, while \texttt{Qwen 3.6 Plus} combines strong TSA with CONFAB. $=0.571 \pm 0.111$. Overall, E1 partially supports the hypothesis: some agents can use reflected evidence to infer who they are, but neither mirror consultation nor self-referential language is sufficient on its own.

Text-based metrics further caution against treating self-referential language as evidence of self-grounding. \texttt{Claude Sonnet 4.6} achieves perfect CAAL but also a relatively high
CR of $0.619 \pm 0.109$, while \texttt{Qwen 3.6 Plus} combines strong TSA with CR $= 0.571 \pm 0.111$. Overall, E1 partially supports the hypothesis: some agents can use reflected evidence to infer who they are, but neither mirror consultation nor self-attribution alone is sufficient to establish grounded self-identification.

\textbf{E2: What Does the Agent Do When the Mirror Is Gone?}
E2 serves as a negative control by removing the mirror while leaving the task otherwise unchanged. Under the hypothesis in Sec.~\ref{Experimental Conditions}, performance should collapse when self-referential visual evidence is unavailable; any residual success should be interpreted as shortcut behavior, prior bias, or guessing rather than grounded self-identification.

The task-success metrics in Table~\ref{tab:all-results} and Figure~\ref{fig:spider_plot} broadly support this interpretation. Most models show low TSA when the mirror is removed: \texttt{Ministral 3 14B} reaches TSA $=0.000 \pm 0.000$, \texttt{Gemma4 31B} reaches $0.048 \pm 0.048$, \texttt{Gemma4 26B} reaches $0.095 \pm 0.066$, and \texttt{Qwen 3.6 Plus} reaches $0.143 \pm 0.078$. This drop relative to E1 supports the causal role of mirror evidence. However, \texttt{Claude Sonnet 4.6} retains TSA $=0.524 \pm 0.112$, and \texttt{Gemini 2.5 Pro} retains TSA $=0.381 \pm 0.109$. Since the hidden body attribute is not visually observable in E2, these successes cannot be interpreted as self-identification and instead reveal residual shortcut or guessing behavior. 

Mirror interaction collapses completely, with MCR $= 0$ for all models and MTATO correspondingly zero or undefined, confirming that the relevant perceptual channel has been removed. More interestingly, TTD reveals two distinct response patterns: some models commit quickly despite lacking evidence, most clearly \texttt{Gemini 2.5 Pro} (TTD $= 15.333 \pm 2.493$, CR $= 0.952 \pm 0.048$), whereas others continue exploring much longer under uncertainty, such as \texttt{GPT 5.1} (TTD $= 98.476 \pm 0.742$). Several agents continue to assert self-color identifications even when they lack perceptual access to the relevant visual evidence. \texttt{Claude Sonnet 4.6} has CAAL $=0.762 \pm 0.095$ but CR. $=1.000 \pm 0.000$, while \texttt{Gemini 2.5 Pro} and \texttt{Gemini 2.5 Flash} reach CR. $=0.952 \pm 0.048$ and $0.905 \pm 0.066$, respectively. Thus, E2 strongly supports the negative-control hypothesis: without the mirror, agents often guess or confabulate, and fluent self-attribution is not evidence of grounded self-identification.

\textbf{E3: Can the Agent Trust Its Eyes Over a False Belief About Itself?}
E3 introduces a misleading textual statement about the agent's body color while preserving access to the mirror. This condition tests whether the agent privileges reflected visual evidence over an incorrect linguistic prior. The results in Table~\ref{tab:all-results} show that this ability is limited and highly model-dependent.

Task success is substantially lower than in the clean mirror condition for several models. The best-performing model is \texttt{GPT 5.1}, with TSA $=0.524 \pm 0.112$, followed by \texttt{Qwen 3.6 Plus} with TSA $=0.429 \pm 0.111$ and \texttt{Gemini 2.5 Pro} with TSA $=0.381 \pm 0.109$. The most striking degradation is observed for \texttt{Claude Sonnet 4.6}, which drops from TSA $=0.905 \pm 0.066$ in E1 to $0.143 \pm 0.078$ in E3. Thus, success in the clean mirror setting does not imply robustness to language--vision conflict.

The mirror-interaction metrics show that this degradation is not due to failure to consult the mirror. Several models have MCR $=1.000 \pm 0.000$, including \texttt{Claude Sonnet 4.6}, \texttt{Gemini 2.5 Flash}, \texttt{GPT 5.1}, and \texttt{Ministral 3 14B}. MTATO is also perfect for all models with defined values. The main failure mode is therefore epistemic rather than exploratory: agents often access the relevant visual evidence but do not reliably resolve the conflict in favor of perception.

The text-based metrics reveal a further action--language dissociation. \texttt{Gemini 2.5 Flash} reaches CAAL $=0.952 \pm 0.048$ but only TSA $=0.333 \pm 0.105$, and \texttt{Ministral 3 14B} reaches CAAL $=0.762 \pm 0.095$ but TSA $=0.238 \pm 0.095$. These results show that correctly stating the self-color does not guarantee a correct embodied action. E3, therefore, only partially supports the hypothesis: agents can sometimes trust visual evidence over a false textual belief, but this ability is unstable and often breaks between verbal attribution and navigation.

\textbf{E4: Can the Agent Tell Which Reflection Is Itself When Others Look Similar?}
E4 probes self--other disambiguation. The hypothesis is that successful agents should go beyond static appearance matching and use action-contingent evidence to identify which reflected body corresponds to themselves. The results in Table~\ref{tab:all-results} and Figure \ref{fig:spider_plot} suggest that some models can handle this ambiguity, but the capability remains fragile.

In terms of task success, \texttt{Gemini 2.5 Pro} performs best, with TSA $=0.714 \pm 0.101$. \texttt{Claude Sonnet 4.6} and \texttt{Qwen 3.6 Plus} both reach TSA $=0.667 \pm 0.105$. These results indicate that self--other ambiguity does not fully prevent mirror-guided action for stronger models. However, the performance gap across models remains large: \texttt{Gemma4 26B} reaches only TSA $=0.190 \pm 0.088$ despite high mirror consultation.

The mirror-interaction metrics again separate evidence seeking from self-identification. MCR is high for all models, and MTATO is generally high, but TSA varies widely. \texttt{Gemma4 26B} has MCR $=0.952 \pm 0.048$ and MTATO $=1.000 \pm 0.000$, but low TSA. \texttt{Ministral 3 14B} has MCR $=1.000 \pm 0.000$ and MTATO $=1.000 \pm 0.000$, but TSA $=0.333 \pm 0.105$. These failures suggest that agents often inspect the mirror before acting, yet fail to determine which reflected entity is causally linked to themselves.

Text-based metrics remain only partially aligned with behavior. Stronger models often achieve high CAAL, including \texttt{Claude Sonnet 4.6} with CAAL $=1.000 \pm 0.000$ and \texttt{Gemini 2.5 Pro} with CAAL $=0.952 \pm 0.048$.  Overall, E4 partially supports the hypothesis: stronger agents can sometimes identify which reflection is themselves, but self--other disambiguation remains brittle and cannot be inferred from mirror consultation alone.

\textbf{E5: Can the Agent Still Recognize Itself Through a Partially Occluded Reflection?}
\vspace{-0.2cm}

E5 tests robustness under degraded mirror evidence. If agents relied on robust causal or temporal cues rather than static visual matching alone, performance should degrade gradually under occlusion. The results in Table~\ref{tab:all-results} instead show a broad collapse, indicating that current mirror-guided self-identification remains brittle.

Task success is substantially reduced across models. \texttt{Claude Sonnet 4.6} remains the strongest model but drops to TSA $=0.476 \pm 0.112$. \texttt{GPT 5.1} follows with TSA $=0.238 \pm 0.095$, while most other models are near the bottom of the range: \texttt{Gemma4 26B} reaches TSA $=0.048 \pm 0.048$, \texttt{Gemma4 31B} reaches $0.095 \pm 0.066$, \texttt{Qwen 3.6 Plus} reaches $0.143 \pm 0.078$, and \texttt{Ministral 3 14B} reaches $0.000 \pm 0.000$.

The mirror-interaction metrics show that this failure is not caused by a lack of evidence seeking. MCR remains high for all models, ranging from $0.810 \pm 0.088$ to $1.000 \pm 0.000$. For instance, \texttt{Ministral 3 14B} has MCR $=1.000 \pm 0.000$, yet obtains TSA $=0.000 \pm 0.000$. Similarly, \texttt{Gemini 2.5 Flash} has perfect MCR, but TSA $=0.190 \pm 0.088$. Hence, the limiting factor is not whether agents look at the mirror, but whether they can interpret incomplete reflected evidence.

The text-based metrics show a distinctive form of failure.  CAAL collapses to zero or near zero for most models, including \texttt{Claude Sonnet 4.6},
\texttt{Gemma4 26B}, \texttt{Gemma4 31B}, and \texttt{GPT 5.1}, while
CR is near zero across almost all models. This should not be interpreted as
improved grounding; rather, many agents fail to make a concrete self-attribution at all under extended uncertainty. E5 therefore demonstrates that mirror access alone is insufficient for robust self-identification in open-ended settings. Although agents continue to seek mirror evidence, they generally fail to use it effectively when the reflection is partially occluded, thereby rejecting the robustness hypothesis.

\begin{table*}[t]
\centering
%\scriptsize
\resizebox{\textwidth}{!}{%
\begin{tabular}{lcccccc}
\toprule
\textbf{Model} & \textbf{TSA} $\uparrow$ & \textbf{TTD} $\downarrow$ & \textbf{MCR} $\uparrow$ & \textbf{MTATO} $\uparrow$ & \textbf{CAAL} $\uparrow$ & \textbf{CR} $\downarrow$ \\
\midrule

\rowcolor{gray!15}
\multicolumn{7}{c}{\textbf{E1: Active mirror-based self-identification}} \\
\midrule
\texttt{Ministral 3 14B} & $0.10 \pm 0.07$ & $94.43 \pm 3.56$ & $0.86 \pm 0.08$ & $\mathbf{1.00 \pm 0.00}$ & $0.67 \pm 0.11$ & $0.24 \pm 0.10$ \\
\texttt{Gemma4 26B} & $0.57 \pm 0.11$ & $93.43 \pm 4.08$ & $\mathbf{1.00 \pm 0.00}$ & $\mathbf{1.00 \pm 0.00}$ & $\underline{0.91 \pm 0.07}$ & $\underline{0.19 \pm 0.09}$ \\
\texttt{Gemma4 31B} & $0.57 \pm 0.11$ & $47.67 \pm 8.15$ & $0.71 \pm 0.10$ & $0.58 \pm 0.15$ & $0.52 \pm 0.11$ & $\mathbf{0.14 \pm 0.08}$ \\
\texttt{Gemini 2.5 Flash} & $0.43 \pm 0.11$ & $73.43 \pm 7.63$ & $0.76 \pm 0.10$ & $0.67 \pm 0.17$ & $0.76 \pm 0.10$ & $0.43 \pm 0.11$ \\
\texttt{Gemini 2.5 Pro} & $0.43 \pm 0.11$ & $\mathbf{24.48 \pm 4.27}$ & $0.76 \pm 0.10$ & $0.78 \pm 0.15$ & $0.76 \pm 0.10$ & $0.24 \pm 0.10$ \\
\texttt{Qwen 3.6 Plus} & $\underline{0.71 \pm 0.10}$ & $\underline{37.10 \pm 6.33}$ & $0.71 \pm 0.10$ & $0.80 \pm 0.11$ & $0.76 \pm 0.10$ & $0.57 \pm 0.11$ \\
\texttt{GPT 5.1} & $0.52 \pm 0.11$ & $58.71 \pm 8.24$ & $0.91 \pm 0.07$ & $\mathbf{1.00 \pm 0.00}$ & $0.81 \pm 0.09$ & $0.33 \pm 0.11$ \\
\texttt{Claude Sonnet 4.6} & $\mathbf{0.91 \pm 0.07}$ & $76.33 \pm 6.24$ & $\underline{0.95 \pm 0.05}$ & $\underline{0.95 \pm 0.05}$ & $\mathbf{1.00 \pm 0.00}$ & $0.62 \pm 0.11$ \\

\midrule
\rowcolor{gray!15}
\multicolumn{7}{c}{\textbf{E2: Self-identification without mirror evidence}} \\
\midrule
\texttt{Ministral 3 14B} & $0.00 \pm 0.00$ & $95.62 \pm 3.06$ & $\mathbf{0.00 \pm 0.00}$ & $\mathbf{0.00 \pm 0.00}$ & $0.10 \pm 0.07$ & $\mathbf{0.24 \pm 0.10}$ \\
\texttt{Gemma4 26B} & $0.10 \pm 0.07$ & $83.43 \pm 7.49$ & $\mathbf{0.00 \pm 0.00}$ & $\mathbf{0.00 \pm 0.00}$ & $0.10 \pm 0.07$ & $0.43 \pm 0.11$ \\
\texttt{Gemma4 31B} & $0.05 \pm 0.05$ & $78.33 \pm 7.10$ & $\mathbf{0.00 \pm 0.00}$ & $\mathbf{0.00 \pm 0.00}$ & $0.05 \pm 0.05$ & $\underline{0.33 \pm 0.11}$ \\
\texttt{Gemini 2.5 Flash} & $0.10 \pm 0.07$ & $94.67 \pm 3.94$ & $\mathbf{0.00 \pm 0.00}$ & $\mathbf{0.00 \pm 0.00}$ & $0.14 \pm 0.08$ & $0.91 \pm 0.07$ \\
\texttt{Gemini 2.5 Pro} & $\underline{0.38 \pm 0.11}$ & $\mathbf{15.33 \pm 2.49}$ & $\mathbf{0.00 \pm 0.00}$ & $\mathbf{0.00 \pm 0.00}$ & $\underline{0.52 \pm 0.11}$ & $0.95 \pm 0.05$ \\
\texttt{Qwen 3.6 Plus} & $0.14 \pm 0.08$ & $\underline{67.67 \pm 8.94}$ & $\mathbf{0.00 \pm 0.00}$ & $\mathbf{0.00 \pm 0.00}$ & $0.14 \pm 0.08$ & $0.76 \pm 0.10$ \\
\texttt{GPT 5.1} & $0.19 \pm 0.09$ & $98.48 \pm 0.74$ & $\mathbf{0.00 \pm 0.00}$ & $\mathbf{0.00 \pm 0.00}$ & $0.00 \pm 0.00$ & $0.38 \pm 0.11$ \\
\texttt{Claude Sonnet 4.6} & $\mathbf{0.52 \pm 0.11}$ & $91.67 \pm 2.55$ & $\mathbf{0.00 \pm 0.00}$ & $\mathbf{0.00 \pm 0.00}$ & $\mathbf{0.76 \pm 0.10}$ & $1.00 \pm 0.00$ \\

\midrule
\rowcolor{gray!15}
\multicolumn{7}{c}{\textbf{E3: Self-identification under conflicting linguistic information}} \\
\midrule
\texttt{Ministral 3 14B} & $0.24 \pm 0.10$ & $84.33 \pm 5.83$ & $\mathbf{1.00 \pm 0.00}$ & $\mathbf{1.00 \pm 0.00}$ & $0.76 \pm 0.10$ & $\mathbf{0.19 \pm 0.09}$ \\
\texttt{Gemma4 26B} & $0.29 \pm 0.10$ & $57.90 \pm 10.07$ & $0.67 \pm 0.11$ & $\mathbf{1.00 \pm 0.00}$ & $0.62 \pm 0.11$ & $\mathbf{0.19 \pm 0.09}$ \\
\texttt{Gemma4 31B} & $0.33 \pm 0.11$ & $\mathbf{20.00 \pm 7.44}$ & $0.57 \pm 0.11$ & $\mathbf{1.00 \pm 0.00}$ & $0.33 \pm 0.11$ & $\mathbf{0.19 \pm 0.09}$ \\
\texttt{Gemini 2.5 Flash} & $0.33 \pm 0.11$ & $73.71 \pm 7.21$ & $\mathbf{1.00 \pm 0.00}$ & $\mathbf{1.00 \pm 0.00}$ & $\mathbf{0.95 \pm 0.05}$ & $0.52 \pm 0.11$ \\
\texttt{Gemini 2.5 Pro} & $0.38 \pm 0.11$ & $\underline{52.81 \pm 9.18}$ & $\underline{0.95 \pm 0.05}$ & $\mathbf{1.00 \pm 0.00}$ & $\underline{0.81 \pm 0.09}$ & $0.52 \pm 0.11$ \\
\texttt{Qwen 3.6 Plus} & $\underline{0.43 \pm 0.11}$ & $57.10 \pm 8.98$ & $0.86 \pm 0.08$ & $\mathbf{1.00 \pm 0.00}$ & $0.71 \pm 0.10$ & $0.48 \pm 0.11$ \\
\texttt{GPT 5.1} & $\mathbf{0.52 \pm 0.11}$ & $57.19 \pm 7.84$ & $\mathbf{1.00 \pm 0.00}$ & $\mathbf{1.00 \pm 0.00}$ & $0.62 \pm 0.11$ & $0.57 \pm 0.11$ \\
\texttt{Claude Sonnet 4.6} & $0.14 \pm 0.08$ & $75.67 \pm 7.57$ & $\mathbf{1.00 \pm 0.00}$ & $\mathbf{1.00 \pm 0.00}$ & $0.29 \pm 0.10$ & $\underline{0.38 \pm 0.11}$ \\

\midrule
\rowcolor{gray!15}
\multicolumn{7}{c}{\textbf{E4: Self--other disambiguation under visual ambiguity}} \\
\midrule
\texttt{Ministral 3 14B} & $0.33 \pm 0.11$ & $80.57 \pm 7.17$ & $\mathbf{1.00 \pm 0.00}$ & $\mathbf{1.00 \pm 0.00}$ & $0.81 \pm 0.09$ & $\underline{0.24 \pm 0.10}$ \\
\texttt{Gemma4 26B} & $0.19 \pm 0.09$ & $87.52 \pm 4.15$ & $\underline{0.95 \pm 0.05}$ & $\mathbf{1.00 \pm 0.00}$ & $0.62 \pm 0.11$ & $0.38 \pm 0.11$ \\
\texttt{Gemma4 31B} & $0.38 \pm 0.11$ & $45.05 \pm 7.32$ & $\mathbf{1.00 \pm 0.00}$ & $\mathbf{1.00 \pm 0.00}$ & $0.48 \pm 0.11$ & $\mathbf{0.14 \pm 0.08}$ \\
\texttt{Gemini 2.5 Flash} & $0.38 \pm 0.11$ & $62.76 \pm 8.46$ & $\underline{0.95 \pm 0.05}$ & $0.88 \pm 0.13$ & $0.86 \pm 0.08$ & $0.38 \pm 0.11$ \\
\texttt{Gemini 2.5 Pro} & $\mathbf{0.71 \pm 0.10}$ & $\underline{42.57 \pm 6.03}$ & $\underline{0.95 \pm 0.05}$ & $\underline{0.93 \pm 0.07}$ & $\underline{0.95 \pm 0.05}$ & $0.48 \pm 0.11$ \\
\texttt{Qwen 3.6 Plus} & $\underline{0.67 \pm 0.11}$ & $\mathbf{40.43 \pm 6.39}$ & $0.86 \pm 0.08$ & $\underline{0.93 \pm 0.07}$ & $0.76 \pm 0.10$ & $0.48 \pm 0.11$ \\
\texttt{GPT 5.1} & $0.38 \pm 0.11$ & $75.29 \pm 6.11$ & $\mathbf{1.00 \pm 0.00}$ & $\mathbf{1.00 \pm 0.00}$ & $0.76 \pm 0.10$ & $\underline{0.24 \pm 0.10}$ \\
\texttt{Claude Sonnet 4.6} & $\underline{0.67 \pm 0.11}$ & $78.67 \pm 6.75$ & $\mathbf{1.00 \pm 0.00}$ & $\mathbf{1.00 \pm 0.00}$ & $\mathbf{1.00 \pm 0.00}$ & $0.48 \pm 0.11$ \\

\midrule
\rowcolor{gray!15}
\multicolumn{7}{c}{\textbf{E5: Robust self-identification under occlusion}} \\
\midrule
\texttt{Ministral 3 14B} & $0.00 \pm 0.00$ & $93.43 \pm 4.14$ & $\mathbf{1.00 \pm 0.00}$ & $\mathbf{0.00 \pm 0.00}$ & $\underline{0.05 \pm 0.05}$ & $\mathbf{0.00 \pm 0.00}$ \\
\texttt{Gemma4 26B} & $0.05 \pm 0.05$ & $84.57 \pm 6.86$ & $0.81 \pm 0.09$ & $\mathbf{1.00 \pm 0.00}$ & $0.00 \pm 0.00$ & $\underline{0.05 \pm 0.05}$ \\
\texttt{Gemma4 31B} & $0.10 \pm 0.07$ & $70.00 \pm 6.93$ & $\mathbf{1.00 \pm 0.00}$ & $\mathbf{1.00 \pm 0.00}$ & $0.00 \pm 0.00$ & $\mathbf{0.00 \pm 0.00}$ \\
\texttt{Gemini 2.5 Flash} & $0.19 \pm 0.09$ & $88.86 \pm 3.90$ & $\mathbf{1.00 \pm 0.00}$ & $\mathbf{1.00 \pm 0.00}$ & $\mathbf{0.19 \pm 0.09}$ & $\mathbf{0.00 \pm 0.00}$ \\
\texttt{Gemini 2.5 Pro} & $0.19 \pm 0.09$ & $\mathbf{33.19 \pm 5.21}$ & $\mathbf{1.00 \pm 0.00}$ & $\mathbf{1.00 \pm 0.00}$ & $\underline{0.05 \pm 0.05}$ & $\mathbf{0.00 \pm 0.00}$ \\
\texttt{Qwen 3.6 Plus} & $0.14 \pm 0.08$ & $75.57 \pm 8.12$ & $\underline{0.95 \pm 0.05}$ & $\mathbf{1.00 \pm 0.00}$ & $\underline{0.05 \pm 0.05}$ & $\mathbf{0.00 \pm 0.00}$ \\
\texttt{GPT 5.1} & $\underline{0.24 \pm 0.10}$ & $99.19 \pm 0.15$ & $\underline{0.95 \pm 0.05}$ & $\mathbf{1.00 \pm 0.00}$ & $0.00 \pm 0.00$ & $\mathbf{0.00 \pm 0.00}$ \\
\texttt{Claude Sonnet 4.6} & $\mathbf{0.48 \pm 0.11}$ & $\underline{62.10 \pm 3.43}$ & $\mathbf{1.00 \pm 0.00}$ & $\mathbf{1.00 \pm 0.00}$ & $0.00 \pm 0.00$ & $\mathbf{0.00 \pm 0.00}$ \\

\bottomrule
\end{tabular}%
}
\caption{\textbf{Experimental results across all defined settings}. Values are mean $\pm$ standard error. Best results are in bold, second-best results are underlined.}
\label{tab:all-results}
\end{table*}

% \section{Conclusions}
% \label{sec:conclusions}
% \vspace{-0.3cm}

% We introduced a controlled benchmark for functional mirror-guided self-identification in embodied VLM agents.
% Our experiments show that mirror-guided self-identification can emerge in current VLM agents, but it remains partial, fragile, and strongly model-dependent. Stronger models achieve high target-selection accuracy in the mirror condition, while weaker models often inspect the mirror without reliably using reflected evidence for action. Removing the mirror substantially reduces performance, confirming that visual self-evidence is causally important. However, mirror consultation alone is not sufficient: agents may look at the mirror in the correct temporal order yet still fail to extract the relevant self-attribute. Language--vision conflict further reveals differences in epistemic grounding, as only some models privilege perceptual evidence over misleading textual cues. Across conditions, self-referential language is not a reliable proxy for grounded self-identification, since models can produce fluent first-person explanations even when mirror evidence is absent, misleading, or unusable.

\section{Conclusions}
\label{sec:conclusions}

We introduced a controlled benchmark for functional mirror-guided
self-identification in embodied VLM agents. Our results show that this
capability can emerge, but it remains partial, fragile, and strongly
model-dependent. Performance is highest in the clean mirror condition, but drops when mirror evidence is removed, when language conflicts with perception, and under self--other ambiguity or cluttered reflected scenes. This supports the causal role of reflected visual evidence while also showing that success in the basic condition should not be overinterpreted as robust self-awareness-like behavior.

Across conditions, the main pattern is a dissociation between accessing
evidence and using it. Many agents consult the mirror, often before acting, yet still fail to extract the relevant self-attribute or translate it into the correct final decision. Language introduces a second dissociation: agents can produce plausible or even correct self-attributions without reliable perceptual grounding, as shown by confabulation in the no-mirror control and by the gap between CAAL and TSA under misleading language. Overall, the strongest evidence
for mirror-grounded self-identification appears only when task success, mirror consultation, temporal ordering, and correct self-attribution align.

\textbf{Limitations and Future Work.} This work should be interpreted as a test of functional mirror-guided self-identification, not as philosophical self-awareness. Moreover, VLMs may still partly rely on pretraining priors about mirrors and reflections.  We mitigate this by using an unlabeled reflective surface and by avoiding an explicit, canonical mirror frame in the environment. This design reduces direct semantic shortcuts, although it cannot fully eliminate learned visual priors about reflective surfaces.%Overall, our results indicate that current VLM agents can use reflected visual evidence in controlled settings, but they do not yet maintain robust, causally grounded self-identification. 
Future work should extend this setting to multi-agent scenarios, collaborative tasks, video-based evaluation, and action-conditioned world models to test whether stronger representations of action consequences improve embodied self-grounding.

\bibliographystyle{plainnat}
\bibliography{bibliography}

@inproceedings{todorov2012mujoco,
  title={MuJoCo: A physics engine for model-based control},
  author={Todorov, Emanuel and Erez, Tom and Tassa, Yuval},
  booktitle={2012 IEEE/RSJ International Conference on Intelligent Robots and Systems},
  pages={5026--5033},
  year={2012},
  organization={IEEE},
  doi={10.1109/IROS.2012.6386109}
}

@article{Brandl2018The,
title={The puzzle of mirror self-recognition},
author={Johannes L. Brandl},
journal={Phenomenology and the Cognitive Sciences},
year={2018},
volume={17},
pages={279-304},
doi={10.1007/s11097-016-9486-7}
}

@article{Chen2024Are,
title={Are We on the Right Way for Evaluating Large Vision-Language Models?},
author={Lin Chen and Jinsong Li and Xiao-wen Dong and Pan Zhang and Yuhang Zang and Zehui Chen and Haodong Duan and Jiaqi Wang and Yu Qiao and Dahua Lin and Feng Zhao},
journal={ArXiv},
year={2024},
volume={abs/2403.20330},
doi={10.48550/arxiv.2403.20330}
}

@article{Hoffmann2020Robot,
title={Robot in the Mirror: Toward an Embodied Computational Model of Mirror Self-Recognition},
author={M. Hoffmann and Shengzhi Wang and Vojtěch Outrata and E. Alzueta and Pablo Lanillos},
journal={KI - Künstliche Intelligenz},
year={2020},
volume={35},
pages={37 - 51},
doi={10.1007/s13218-020-00701-7}
}

@article{Li2025Self-Rewarding,
title={Self-Rewarding Vision-Language Model via Reasoning Decomposition},
author={Zongxia Li and Wenhao Yu and Chengsong Huang and Rui Liu and Zhenwen Liang and Fuxiao Liu and Jingxi Che and Dian Yu and J. Boyd-Graber and Haitao Mi and Dong Yu},
journal={ArXiv},
year={2025},
volume={abs/2508.19652},
doi={10.48550/arxiv.2508.19652}
}

@article{Li2024NaturalBench:,
title={NaturalBench: Evaluating Vision-Language Models on Natural Adversarial Samples},
author={Baiqi Li and Zhiqiu Lin and Wenxuan Peng and Jean de Dieu Nyandwi and Daniel Jiang and Zixian Ma and Simran Khanuja and Ranjay Krishna and Graham Neubig and Deva Ramanan},
journal={ArXiv},
year={2024},
volume={abs/2410.14669},
doi={10.48550/arxiv.2410.14669}
}

@article{Liu2023MMBench:,
title={MMBench: Is Your Multi-modal Model an All-around Player?},
author={Yuanzhan Liu and Haodong Duan and Yuanhan Zhang and Bo Li and Songyang Zhang and Wangbo Zhao and Yike Yuan and Conghui He and Ziwei Liu and Kai Chen and Dahua Lin},
journal={ArXiv},
year={2023},
volume={abs/2307.06281},
doi={10.48550/arxiv.2307.06281}
}

@article{Mitchell1993Mental,
title={Mental models of mirror-self-recognition: Two theories},
author={R. Mitchell},
journal={New Ideas in Psychology},
year={1993},
volume={11},
pages={295-325},
doi={10.1016/0732-118x(93)90002-u}
}

@article{Pipitone2021Robot,
title={Robot passes the mirror test by inner speech},
author={A. Pipitone and A. Chella},
journal={Robotics Auton. Syst.},
year={2021},
volume={144},
pages={103838},
doi={10.1016/j.robot.2021.103838}
}

@inproceedings{hu2019you,
  title={Are you looking? grounding to multiple modalities in vision-and-language navigation},
  author={Hu, Ronghang and Fried, Daniel and Rohrbach, Anna and Klein, Dan and Darrell, Trevor and Saenko, Kate},
  booktitle={Proceedings of the 57th Annual Meeting of the Association for Computational Linguistics},
  pages={6551--6557},
  year={2019}
}

@article{Rahmanzadehgervi2024Vision,
title={Vision language models are blind},
author={Pooyan Rahmanzadehgervi and Logan Bolton and Mohammad Reza Taesiri and A. Nguyen},
journal={ArXiv},
year={2024},
volume={abs/2407.06581},
doi={10.48550/arxiv.2407.06581}
}

@article{Shimada2022Multisensory,
title={Multisensory and Sensorimotor Integration in the Embodied Self: Relationship between Self-Body Recognition and the Mirror Neuron System},
author={Sotaro Shimada},
journal={Sensors (Basel, Switzerland)},
year={2022},
volume={22},
doi={10.3390/s22135059}
}

@article{steels2008robot,
  title={The robot in the mirror},
  author={Steels, Luc and Spranger, Michael},
  journal={Connection Science},
  volume={20},
  number={4},
  pages={337--358},
  year={2008},
  publisher={Taylor \& Francis}
}

@article{Soderlund2022When,
title={When service robots look at themselves in the mirror: An examination of the effects of perceptions of robotic self-recognition},
author={Magnus Söderlund},
journal={Journal of Retailing and Consumer Services},
year={2022},
doi={10.1016/j.jretconser.2021.102820}
}

@inproceedings{wang2025enhancing,
  title={Enhancing visual-language modality alignment in large vision language models via self-improvement},
  author={Wang, Xiyao and Chen, Jiuhai and Wang, Zhaoyang and Zhou, Yuhang and Zhou, Yiyang and Yao, Huaxiu and Zhou, Tianyi and Goldstein, Tom and Bhatia, Parminder and Kass-Hout, Taha and others},
  booktitle={Findings of the Association for Computational Linguistics: NAACL 2025},
  pages={268--282},
  year={2025}
}

@article{yang2019embodied,
  title={Embodied visual recognition},
  author={Yang, Jianwei and Ren, Zhile and Xu, Mingze and Chen, Xinlei and Crandall, David and Parikh, Devi and Batra, Dhruv},
  journal={arXiv preprint arXiv:1904.04404},
  year={2019}
}

@article{Wang2024Enhancing,
title={Enhancing Visual-Language Modality Alignment in Large Vision Language Models via Self-Improvement},
author={Xiyao Wang and Jiuhai Chen and Zhaoyang Wang and Yuhang Zhou and Yiyang Zhou and Huaxiu Yao and Tianyi Zhou and Tom Goldstein and Parminder Bhatia and Furong Huang and Cao Xiao},
journal={ArXiv},
year={2024},
volume={abs/2405.15973},
doi={10.48550/arxiv.2405.15973}
}

@article{xu2024lvlm,
  title={Lvlm-ehub: A comprehensive evaluation benchmark for large vision-language models},
  author={Xu, Peng and Shao, Wenqi and Zhang, Kaipeng and Gao, Peng and Liu, Shuo and Lei, Meng and Meng, Fanqing and Huang, Siyuan and Qiao, Yu and Luo, Ping},
  journal={IEEE Transactions on Pattern Analysis and Machine Intelligence},
  volume={47},
  number={3},
  pages={1877--1893},
  year={2024},
  publisher={IEEE}
}

@inproceedings{Yue2024SC-,
title={SC- Tune: Unleashing Self-Consistent Referential Comprehension in Large Vision Language Models},
author={Tongtian Yue and Jie Cheng and Longteng Guo and Xingyuan Dai and Zijia Zhao and Xingjian He and Gang Xiong and Yisheng Lv and Jing Liu},
booktitle={2024 IEEE/CVF Conference on Computer Vision and Pattern Recognition (CVPR)},
year={2024},
pages={13073-13083},
doi={10.1109/cvpr52733.2024.01242}
}

@article{Zeng2017Toward,
title={Toward Robot Self-Consciousness (II): Brain-Inspired Robot Bodily Self Model for Self-Recognition},
author={Yi Zeng and Yuxuan Zhao and Jun Bai and Bo Xu},
journal={Cognitive Computation},
year={2017},
volume={10},
pages={307 - 320},
doi={10.1007/s12559-017-9505-1}
}

@article{Zhang2023Vision-Language,
title={Vision-Language Models for Vision Tasks: A Survey},
author={Jingyi Zhang and Jiaxing Huang and Sheng Jin and Shijian Lu},
journal={IEEE Transactions on Pattern Analysis and Machine Intelligence},
year={2023},
volume={46},
pages={5625-5644},
doi={10.1109/tpami.2024.3369699}
}

@article{zhou2015monkeys,
  title={Monkeys pass the mirror test after training},
  author={Zhou, Wen and Jiang, Yi},
  journal={Science China. Life Sciences},
  volume={58},
  number={4},
  pages={405},
  year={2015},
  publisher={Springer Nature BV}
}

@article{Zhou2024Calibrated,
title={Calibrated Self-Rewarding Vision Language Models},
author={Yiyang Zhou and Zhiyuan Fan and Dongjie Cheng and Sihan Yang and Zhaorun Chen and Chenhang Cui and Xiyao Wang and Yun Li and Linjun Zhang and Huaxiu Yao},
journal={ArXiv},
year={2024},
volume={abs/2405.14622},
doi={10.48550/arxiv.2405.14622}
}

@article{gallup1982selfawareness,
  author  = {Gallup, Gordon G.},
  title   = {Self-awareness and the emergence of mind in primates},
  journal = {American Journal of Primatology},
  volume  = {2},
  pages   = {237--248},
  year    = {1982}
}

@incollection{gallup2002mirrortest,
  author    = {Gallup, Gordon G. and Anderson, James R. and Shillito, Diane J.},
  title     = {The mirror test},
  booktitle = {The Cognitive Animal: Empirical and Theoretical Perspectives on Animal Cognition},
  editor    = {Bekoff, Marc and Allen, Colin and Burghardt, Gordon M.},
  publisher = {MIT Press},
  address   = {Cambridge, MA},
  year      = {2002}
}

@article{bekoff2004reflections,
  author  = {Bekoff, Marc and Sherman, Paul W.},
  title   = {Reflections on animal selves},
  journal = {Trends in Ecology \& Evolution},
  volume  = {19},
  number  = {4},
  pages   = {176--180},
  year    = {2004}
}

@article{plotnik2006selfrecognition,
  author  = {Plotnik, Joshua M. and de Waal, Frans B. M. and Reiss, Diana},
  title   = {Self-recognition in an Asian elephant},
  journal = {Proceedings of the National Academy of Sciences},
  volume  = {103},
  number  = {45},
  pages   = {17053--17057},
  year    = {2006}
}

@article{prior2008magpie,
  author  = {Prior, Helmut and Schwarz, Ariane and G{\"u}nt{\"u}rk{\"u}n, Onur},
  title   = {Mirror-induced behavior in the magpie ({Pica pica}): Evidence of self-recognition},
  journal = {PLoS Biology},
  volume  = {6},
  number  = {8},
  pages   = {e202},
  year    = {2008}
}

@misc{marten1994dolphin,
  author       = {Marten, Ken and Psarakos, Suchi},
  title        = {Evidence of Self-Awareness in the Bottlenose Dolphin ({Tursiops truncatus})},
  publisher    = {Cambridge University Press},
  year         = {1994},
}

@article{suarez1981chimpanzees,
  author  = {Suarez, Susan D. and Gallup, Gordon G.},
  title   = {Self-recognition in chimpanzees and orangutans, but not gorillas},
  journal = {Journal of Human Evolution},
  volume  = {10},
  number  = {2},
  pages   = {175--188},
  year    = {1981}
}

@article{mitchell1997kinesthetic,
  author  = {Mitchell, Robert W.},
  title   = {Kinesthetic-visual matching and the self-concept as explanations of mirror-self-recognition},
  journal = {Journal for the Theory of Social Behaviour},
  volume  = {27},
  number  = {1},
  year    = {1997}
}

@article{gallup1970chimpanzees,
  author  = {Gallup, Gordon G.},
  title   = {Chimpanzees: Self-Recognition},
  journal = {Science},
  volume  = {167},
  number  = {3914},
  pages   = {86--87},
  year    = {1970},
  doi     = {10.1126/science.167.3914.86}
}

@article{morin1989gallupsmirrors,
  author  = {Morin, Alain and DeBlois, Sandra},
  title   = {Gallup's Mirrors: More Than an Operationalization of Self-Awareness in Primates?},
  journal = {Psychological Reports},
  volume  = {65},
  number  = {1},
  pages   = {287--291},
  year    = {1989},
  doi     = {10.2466/pr0.1989.65.1.287}
}

@article{morrison2018precocious,
  title={Precocious development of self-awareness in dolphins},
  author={Morrison, Rachel and Reiss, Diana},
  journal={PLoS One},
  volume={13},
  number={1},
  pages={e0189813},
  year={2018},
  publisher={Public Library of Science San Francisco, CA USA}
}

@article{chang2017monkeymsr,
  author  = {Chang, Liangtang and Zhang, Shikun and Poo, Mu-ming and Gong, Neng},
  title   = {Spontaneous expression of mirror self-recognition in monkeys after learning precise visual-proprioceptive association for mirror images},
  journal = {Proceedings of the National Academy of Sciences},
  volume  = {114},
  number  = {12},
  pages   = {3258--3263},
  year    = {2017},
  doi     = {10.1073/pnas.1620764114}
}

@misc{singh2026openaigpt5card,
      title={OpenAI GPT-5 System Card}, 
      author={Aaditya Singh et. al},
      year={2026},
      eprint={2601.03267},
      archivePrefix={arXiv},
      primaryClass={cs.CL},
      url={https://arxiv.org/abs/2601.03267}, 
}

@article{qwen3technicalreport2025,
  title={Qwen3 Technical Report},
  author={Qwen Team},
  journal={arXiv preprint arXiv:2505.09388},
  year={2025},
  url={https://arxiv.org/abs/2505.09388}
}

@article{Comanici2025Gemini2P,
  title={Gemini 2.5: Pushing the Frontier with Advanced Reasoning, Multimodality, Long Context, and Next Generation Agentic Capabilities},
  author={Comanici, Gheorghe and others},
  journal={arXiv preprint arXiv:2507.06261},
  year={2025},
  url={https://arxiv.org/abs/2507.06261}
}

@misc{anthropic2026claude46,
  author       = {Anthropic},
  title        = {Introducing {Claude} {Sonnet} 4.6},
  year         = {2026},
  month        = {February},
  url          = {https://www.anthropic.com/news/claude-sonnet-4-6},
  note         = {Accessed: 2026-05-07}
}

@article{liu2026ministral,
  title={Ministral 3},
  author={Liu, Alexander H and Khandelwal, Kartik and Subramanian, Sandeep and Jouault, Victor and Rastogi, Abhinav and Sad{\'e}, Adrien and Jeffares, Alan and Jiang, Albert and Cahill, Alexandre and Gavaudan, Alexandre and others},
  journal={arXiv preprint arXiv:2601.08584},
  year={2026}
}

\newpage
\clearpage
\FloatBarrier
\appendix

\section{Additional Results}
\label{sec:appendix}

\subsection{Auxiliary Metrics and Relative Results}
\label{sec:appendix-aux-metrics}

In addition to the core metrics reported in the main paper, we retain four
auxiliary metrics that provide additional diagnostic information about
completion policy, persistence of mirror exposure, and belief revision.
These metrics were excluded from the main text because they are either
partially conditioned or secondary to the core task-level measures, but they
remain useful in the supplementary analyses below.

Using the notation introduced in Sec.~\ref{sec:appendix-eval}, let $T_i$ denote
the length of episode $i$, $\tau_i$ the first timestep at which the agent
outputs \texttt{done} when such a timestep exists, $c_i^\star$ the ground-truth
body color, $\hat{c}_{i,t}$ the self-color reported at timestep $t$, and
$m_{i,t} \in \{0,1\}$ whether mirror evidence is visible at timestep $t$.

\paragraph{Completion-Conditioned Task Success.}
Let $C_i = \mathbf{1}[\tau_i \text{ exists}]$ denote the indicator that episode
$i$ terminates with \texttt{done}. The completion-conditioned success rate is
\[
\mathrm{TSA}_{\mathrm{complete}}
=
\frac{\sum_i C_i \, \mathrm{TSA}_i}{\sum_i C_i},
\]
and is reported in tables as \texttt{TSA-C}. This quantity separates incorrect
final decisions from failure to terminate appropriately.

\paragraph{Mirror Gaze Duration.}
Mirror Gaze Duration is defined at the episode level as
\[
\mathrm{MGD}_i = \sum_{t=1}^{T_i} m_{i,t},
\]
that is, the number of timesteps in which mirror evidence is visible during the
trajectory. It measures persistence of mirror exposure rather than successful
use of that exposure.

\paragraph{Self-Correction.}
Let
\[
t_i^{\mathrm{mirror}} = \min \{t : m_{i,t}=1\}
\]
denote the first timestep at which mirror evidence becomes available, whenever
such a timestep exists. Let
\[
t_i^{\mathrm{guess}}
=
\min \{t < t_i^{\mathrm{mirror}} : \hat{c}_{i,t} \notin \{\texttt{unknown}, \texttt{null}, \varnothing\}\}
\]
denote the first committed pre-mirror self-attribution, when it exists. The
self-correction applicability indicator is
\[
A_i
=
\mathbf{1}\!\left[
t_i^{\mathrm{mirror}} \text{ exists},\;
t_i^{\mathrm{guess}} \text{ exists},\;
\hat{c}_{i,t_i^{\mathrm{guess}}} \ne c_i^\star
\right].
\]
For applicable episodes, self-correction is defined as
\[
\mathrm{SC}_i
=
\mathbf{1}\!\left[
\exists t \in [t_i^{\mathrm{mirror}}, \tilde{\tau}_i]
\text{ such that }
\hat{c}_{i,t} = c_i^\star
\right],
\]
where $\tilde{\tau}_i = \tau_i$ if the episode terminates and
$\tilde{\tau}_i = T_i$ otherwise.

\paragraph{Applicability Rate.}
Because self-correction is meaningful only for episodes in which a wrong
pre-mirror self-attribution is made, we also report the self-correction
applicability rate
\[
\mathrm{AR}_{\mathrm{SC}}
=
\frac{1}{N}\sum_i A_i,
\]
together with the correction rate
\[
\mathrm{SC}
=
\frac{\sum_i A_i \, \mathrm{SC}_i}{\sum_i A_i}.
\]
Thus, $\mathrm{AR}_{\mathrm{SC}}$ measures how often the model enters a state
where correction is possible, whereas $\mathrm{SC}$ measures how often it
successfully revises the initial wrong self-attribution once mirror evidence
becomes available.

Table~\ref{tab:appendix-extended-metrics} reports an extended view of the
results, combining selected core metrics with the auxiliary metrics introduced above. The table is organized to highlight completion policy, persistence of mirror exposure, revision behavior, and language-side diagnostics.

\begin{table*}[t]
\centering
%\scriptsize
\resizebox{\textwidth}{!}{%
\begin{tabular}{lccccccc}
\toprule
\textbf{Model} & \textbf{TSA} $\uparrow$ & \textbf{TSA-C} $\uparrow$ & \textbf{MCR} $\uparrow$ & \textbf{MGD} $\uparrow$ & \textbf{CAAL} $\uparrow$ & \textbf{SC} $\uparrow$ & \textbf{AR$_{\mathrm{SC}}$} $\uparrow$ \\
\midrule

\rowcolor{gray!15}
\multicolumn{8}{c}{\textbf{E1: Active mirror-based self-identification}} \\
\midrule
\texttt{Ministral 3 14B} & $0.10 \pm 0.07$ & $0.67 \pm 0.33$ & $0.86 \pm 0.08$ & $11.81 \pm 2.79$ & $0.67 \pm 0.11$ & $\mathbf{1.00 \pm 0.00}$ & $0.10 \pm 0.06$ \\
\texttt{Gemma4 26B} & $0.57 \pm 0.11$ & $\underline{0.80 \pm 0.11}$ & $\mathbf{1.00 \pm 0.00}$ & $\mathbf{28.48 \pm 4.53}$ & $\underline{0.90 \pm 0.07}$ & $\mathbf{1.00 \pm 0.00}$ & $\underline{0.14 \pm 0.08}$ \\
\texttt{Gemma4 31B} & $0.57 \pm 0.11$ & $\underline{0.80 \pm 0.11}$ & $0.71 \pm 0.10$ & $7.48 \pm 2.14$ & $0.52 \pm 0.11$ & -- & $0.00 \pm 0.00$ \\
\texttt{Gemini 2.5 Flash} & $0.43 \pm 0.11$ & $0.60 \pm 0.13$ & $0.76 \pm 0.10$ & $10.76 \pm 2.41$ & $0.76 \pm 0.10$ & $\mathbf{1.00 \pm 0.00}$ & $0.05 \pm 0.05$ \\
\texttt{Gemini 2.5 Pro} & $0.43 \pm 0.11$ & $0.43 \pm 0.11$ & $0.76 \pm 0.10$ & $6.86 \pm 1.59$ & $0.76 \pm 0.10$ & -- & $0.00 \pm 0.00$ \\
\texttt{Qwen 3.6 Plus} & $\underline{0.71 \pm 0.10}$ & $0.79 \pm 0.10$ & $0.71 \pm 0.10$ & $6.57 \pm 1.62$ & $0.76 \pm 0.10$ & $\underline{0.75 \pm 0.25}$ & $\mathbf{0.19 \pm 0.09}$ \\
\texttt{GPT 5.1} & $0.52 \pm 0.11$ & $0.52 \pm 0.11$ & $0.90 \pm 0.07$ & $12.00 \pm 2.30$ & $0.81 \pm 0.09$ & $0.00 \pm 0.00$ & $0.05 \pm 0.05$ \\
\texttt{Claude Sonnet 4.6} & $\mathbf{0.90 \pm 0.07}$ & $\mathbf{0.90 \pm 0.07}$ & $\underline{0.95 \pm 0.05}$ & $\underline{21.67 \pm 4.55}$ & $\mathbf{1.00 \pm 0.00}$ & $\mathbf{1.00 \pm 0.00}$ & $\underline{0.14 \pm 0.08}$ \\

\midrule
\rowcolor{gray!15}
\multicolumn{8}{c}{\textbf{E2: Self-identification without mirror evidence}} \\
\midrule
\texttt{Ministral 3 14B} & $0.00 \pm 0.00$ & $0.00 \pm 0.00$ & $\mathbf{0.00 \pm 0.00}$ & $\mathbf{0.00 \pm 0.00}$ & $0.10 \pm 0.07$ & -- & $\mathbf{0.00 \pm 0.00}$ \\
\texttt{Gemma4 26B} & $0.10 \pm 0.07$ & $0.29 \pm 0.18$ & $\mathbf{0.00 \pm 0.00}$ & $\mathbf{0.00 \pm 0.00}$ & $0.10 \pm 0.07$ & -- & $\mathbf{0.00 \pm 0.00}$ \\
\texttt{Gemma4 31B} & $0.05 \pm 0.05$ & $0.25 \pm 0.25$ & $\mathbf{0.00 \pm 0.00}$ & $\mathbf{0.00 \pm 0.00}$ & $0.05 \pm 0.05$ & -- & $\mathbf{0.00 \pm 0.00}$ \\
\texttt{Gemini 2.5 Flash} & $0.10 \pm 0.07$ & $0.15 \pm 0.10$ & $\mathbf{0.00 \pm 0.00}$ & $\mathbf{0.00 \pm 0.00}$ & $0.14 \pm 0.08$ & -- & $\mathbf{0.00 \pm 0.00}$ \\
\texttt{Gemini 2.5 Pro} & $\underline{0.38 \pm 0.11}$ & $\underline{0.40 \pm 0.11}$ & $\mathbf{0.00 \pm 0.00}$ & $\mathbf{0.00 \pm 0.00}$ & $\underline{0.52 \pm 0.11}$ & -- & $\mathbf{0.00 \pm 0.00}$ \\
\texttt{Qwen 3.6 Plus} & $0.14 \pm 0.08$ & $0.33 \pm 0.17$ & $\mathbf{0.00 \pm 0.00}$ & $\mathbf{0.00 \pm 0.00}$ & $0.14 \pm 0.08$ & -- & $\mathbf{0.00 \pm 0.00}$ \\
\texttt{GPT 5.1} & $0.19 \pm 0.09$ & $0.19 \pm 0.09$ & $\mathbf{0.00 \pm 0.00}$ & $\mathbf{0.00 \pm 0.00}$ & $0.00 \pm 0.00$ & -- & $\mathbf{0.00 \pm 0.00}$ \\
\texttt{Claude Sonnet 4.6} & $\mathbf{0.52 \pm 0.11}$ & $\mathbf{0.52 \pm 0.11}$ & $\mathbf{0.00 \pm 0.00}$ & $\mathbf{0.00 \pm 0.00}$ & $\mathbf{0.76 \pm 0.10}$ & -- & $\mathbf{0.00 \pm 0.00}$ \\

\midrule
\rowcolor{gray!15}
\multicolumn{8}{c}{\textbf{E3: Self-identification under conflicting linguistic information}} \\
\midrule
\texttt{Ministral 3 14B} & $0.24 \pm 0.10$ & $\mathbf{0.83 \pm 0.17}$ & $\mathbf{1.00 \pm 0.00}$ & $\mathbf{38.81 \pm 3.06}$ & $0.76 \pm 0.10$ & $0.67 \pm 0.33$ & $0.14 \pm 0.08$ \\
\texttt{Gemma4 26B} & $0.29 \pm 0.10$ & $0.60 \pm 0.16$ & $0.67 \pm 0.11$ & $23.05 \pm 4.49$ & $0.62 \pm 0.11$ & $\mathbf{1.00 \pm 0.00}$ & $0.19 \pm 0.09$ \\
\texttt{Gemma4 31B} & $0.33 \pm 0.11$ & $0.78 \pm 0.15$ & $0.57 \pm 0.11$ & $9.67 \pm 3.26$ & $0.33 \pm 0.11$ & $0.67 \pm 0.33$ & $0.14 \pm 0.08$ \\
\texttt{Gemini 2.5 Flash} & $0.33 \pm 0.11$ & $0.47 \pm 0.13$ & $\mathbf{1.00 \pm 0.00}$ & $\underline{38.24 \pm 4.00}$ & $\mathbf{0.95 \pm 0.05}$ & $\mathbf{1.00 \pm 0.00}$ & $\underline{0.52 \pm 0.11}$ \\
\texttt{Gemini 2.5 Pro} & $0.38 \pm 0.11$ & $0.62 \pm 0.14$ & $\underline{0.95 \pm 0.05}$ & $33.33 \pm 5.95$ & $\underline{0.81 \pm 0.09}$ & $\underline{0.80 \pm 0.13}$ & $0.48 \pm 0.11$ \\
\texttt{Qwen 3.6 Plus} & $\underline{0.43 \pm 0.11}$ & $\underline{0.82 \pm 0.12}$ & $0.86 \pm 0.08$ & $32.76 \pm 5.55$ & $0.71 \pm 0.10$ & $\underline{0.80 \pm 0.13}$ & $0.48 \pm 0.11$ \\
\texttt{GPT 5.1} & $\mathbf{0.52 \pm 0.11}$ & $0.52 \pm 0.11$ & $\mathbf{1.00 \pm 0.00}$ & $27.81 \pm 4.02$ & $0.62 \pm 0.11$ & $0.50 \pm 0.15$ & $\mathbf{0.57 \pm 0.11}$ \\
\texttt{Claude Sonnet 4.6} & $0.14 \pm 0.08$ & $0.15 \pm 0.08$ & $\mathbf{1.00 \pm 0.00}$ & $36.57 \pm 4.36$ & $0.29 \pm 0.10$ & $0.12 \pm 0.12$ & $0.38 \pm 0.11$ \\

\midrule
\rowcolor{gray!15}
\multicolumn{8}{c}{\textbf{E4: Self--other disambiguation under visual ambiguity}} \\
\midrule
\texttt{Ministral 3 14B} & $0.33 \pm 0.11$ & $\mathbf{0.88 \pm 0.12}$ & $\mathbf{1.00 \pm 0.00}$ & $\underline{24.81 \pm 3.01}$ & $0.81 \pm 0.09$ & $\mathbf{1.00 \pm 0.00}$ & $0.24 \pm 0.09$ \\
\texttt{Gemma4 26B} & $0.19 \pm 0.09$ & $0.27 \pm 0.12$ & $\underline{0.95 \pm 0.05}$ & $20.81 \pm 3.65$ & $0.62 \pm 0.11$ & $0.67 \pm 0.21$ & $0.29 \pm 0.10$ \\
\texttt{Gemma4 31B} & $0.38 \pm 0.11$ & $0.50 \pm 0.13$ & $\mathbf{1.00 \pm 0.00}$ & $9.86 \pm 1.47$ & $0.48 \pm 0.11$ & $0.00 \pm 0.00$ & $0.14 \pm 0.08$ \\
\texttt{Gemini 2.5 Flash} & $0.38 \pm 0.11$ & $0.53 \pm 0.13$ & $\underline{0.95 \pm 0.05}$ & $21.57 \pm 3.50$ & $0.86 \pm 0.08$ & $0.71 \pm 0.18$ & $0.33 \pm 0.10$ \\
\texttt{Gemini 2.5 Pro} & $\mathbf{0.71 \pm 0.10}$ & $0.71 \pm 0.10$ & $\underline{0.95 \pm 0.05}$ & $23.33 \pm 3.70$ & $\underline{0.95 \pm 0.05}$ & $\underline{0.88 \pm 0.12}$ & $\underline{0.38 \pm 0.11}$ \\
\texttt{Qwen 3.6 Plus} & $\underline{0.67 \pm 0.11}$ & $\underline{0.78 \pm 0.10}$ & $0.86 \pm 0.08$ & $12.43 \pm 2.46$ & $0.76 \pm 0.10$ & $\underline{0.88 \pm 0.12}$ & $\underline{0.38 \pm 0.11}$ \\
\texttt{GPT 5.1} & $0.38 \pm 0.11$ & $0.38 \pm 0.11$ & $\mathbf{1.00 \pm 0.00}$ & $19.29 \pm 2.36$ & $0.76 \pm 0.10$ & $0.80 \pm 0.20$ & $0.24 \pm 0.09$ \\
\texttt{Claude Sonnet 4.6} & $\underline{0.67 \pm 0.11}$ & $0.70 \pm 0.11$ & $\mathbf{1.00 \pm 0.00}$ & $\mathbf{39.14 \pm 4.91}$ & $\mathbf{1.00 \pm 0.00}$ & $\mathbf{1.00 \pm 0.00}$ & $\mathbf{0.43 \pm 0.11}$ \\

\midrule
\rowcolor{gray!15}
\multicolumn{8}{c}{\textbf{E5: Robust self-identification under occlusion}} \\
\midrule
\texttt{Ministral 3 14B} & $0.00 \pm 0.00$ & $0.00 \pm 0.00$ & $\mathbf{1.00 \pm 0.00}$ & $\mathbf{37.19 \pm 3.25}$ & $\underline{0.05 \pm 0.05}$ & -- & $\mathbf{0.00 \pm 0.00}$ \\
\texttt{Gemma4 26B} & $0.05 \pm 0.05$ & $0.06 \pm 0.06$ & $0.81 \pm 0.09$ & $17.43 \pm 4.19$ & $0.00 \pm 0.00$ & -- & $\mathbf{0.00 \pm 0.00}$ \\
\texttt{Gemma4 31B} & $0.10 \pm 0.07$ & $0.15 \pm 0.10$ & $\mathbf{1.00 \pm 0.00}$ & $22.62 \pm 3.97$ & $0.00 \pm 0.00$ & -- & $\mathbf{0.00 \pm 0.00}$ \\
\texttt{Gemini 2.5 Flash} & $0.19 \pm 0.09$ & $0.19 \pm 0.09$ & $\mathbf{1.00 \pm 0.00}$ & $\underline{34.90 \pm 2.83}$ & $\mathbf{0.19 \pm 0.09}$ & -- & $\mathbf{0.00 \pm 0.00}$ \\
\texttt{Gemini 2.5 Pro} & $0.19 \pm 0.09$ & $0.19 \pm 0.09$ & $\mathbf{1.00 \pm 0.00}$ & $17.86 \pm 2.75$ & $\underline{0.05 \pm 0.05}$ & -- & $\mathbf{0.00 \pm 0.00}$ \\
\texttt{Qwen 3.6 Plus} & $0.14 \pm 0.08$ & $\underline{0.38 \pm 0.18}$ & $\underline{0.95 \pm 0.05}$ & $29.38 \pm 5.15$ & $\underline{0.05 \pm 0.05}$ & -- & $\mathbf{0.00 \pm 0.00}$ \\
\texttt{GPT 5.1} & $\underline{0.24 \pm 0.10}$ & $0.24 \pm 0.10$ & $\underline{0.95 \pm 0.05}$ & $27.38 \pm 3.37$ & $0.00 \pm 0.00$ & -- & $\mathbf{0.00 \pm 0.00}$ \\
\texttt{Claude Sonnet 4.6} & $\mathbf{0.48 \pm 0.11}$ & $\mathbf{0.48 \pm 0.11}$ & $\mathbf{1.00 \pm 0.00}$ & $31.48 \pm 2.27$ & $0.00 \pm 0.00$ & -- & $\mathbf{0.00 \pm 0.00}$ \\

\bottomrule
\end{tabular}%
}
\caption{\textbf{Extended appendix metrics across all defined settings}. Values are reported as mean $\pm$ standard error. Best results are shown in bold and second-best results are underlined within each experiment and metric. TSA-C denotes completion-conditioned task success. MGD denotes mirror gaze duration. SC denotes the self-correction rate conditioned on episodes where a wrong pre-mirror self-attribution is made and later mirror evidence becomes available; it is reported as -- when this condition is never met. AR$_{\mathrm{SC}}$ denotes the corresponding applicability rate.}
\label{tab:appendix-extended-metrics}
\end{table*}

\FloatBarrier
\clearpage

\subsection{Additional Analysis}
The main paper reports the complete quantitative results in
Table~\ref{tab:all-results}. Here, we provide complementary visual analyses that
highlight cross-experiment structure, action--attribution dissociations, and
the distinct failure modes of $E_5$.

\begin{figure*}[ht]
    \centering
    \includegraphics[width=1\textwidth]{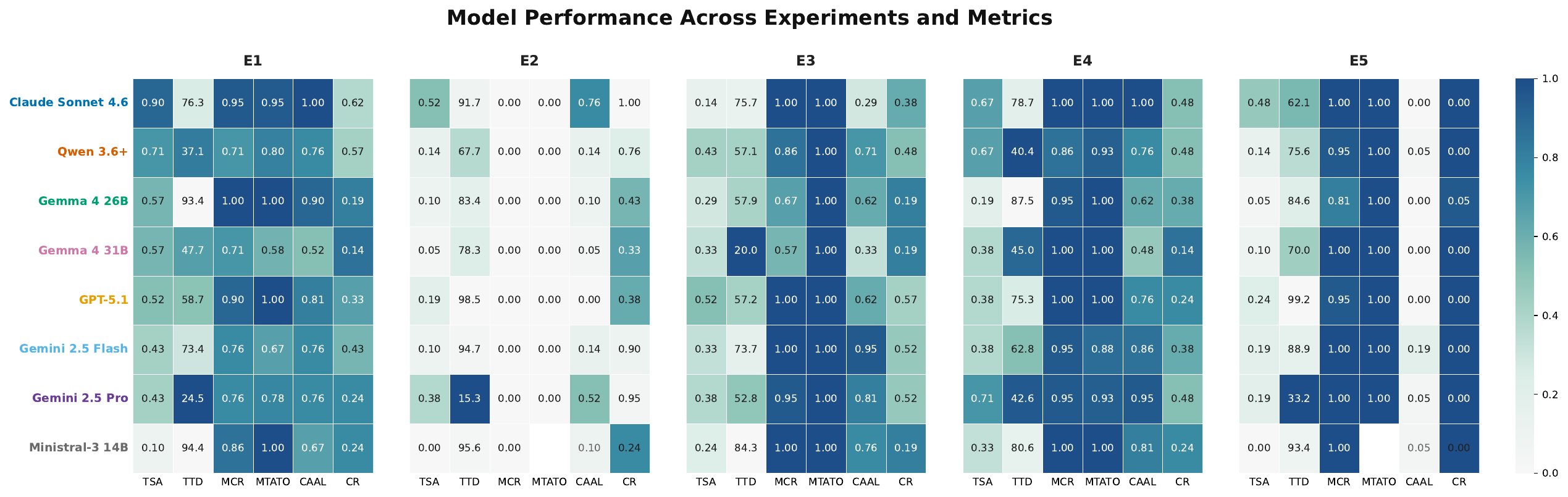}
    \caption{\textbf{Model performance across experiments and metrics.}
    Cell colors encode normalized performance across the reported metric set.
    TTD is reversed and normalized within each experiment, and CR is shown as
    $1-\mathrm{CR}$ so that higher values consistently indicate better
    performance. Cell annotations report raw metric means.}
    \label{fig:appendix-heatmap}
\end{figure*}

Figure~\ref{fig:appendix-heatmap} provides a compact summary of all reported
results. It highlights both the strongest overall models and the changing
metric profiles across experimental conditions.

\begin{figure*}[ht]
    \centering
    \includegraphics[width=1\textwidth]{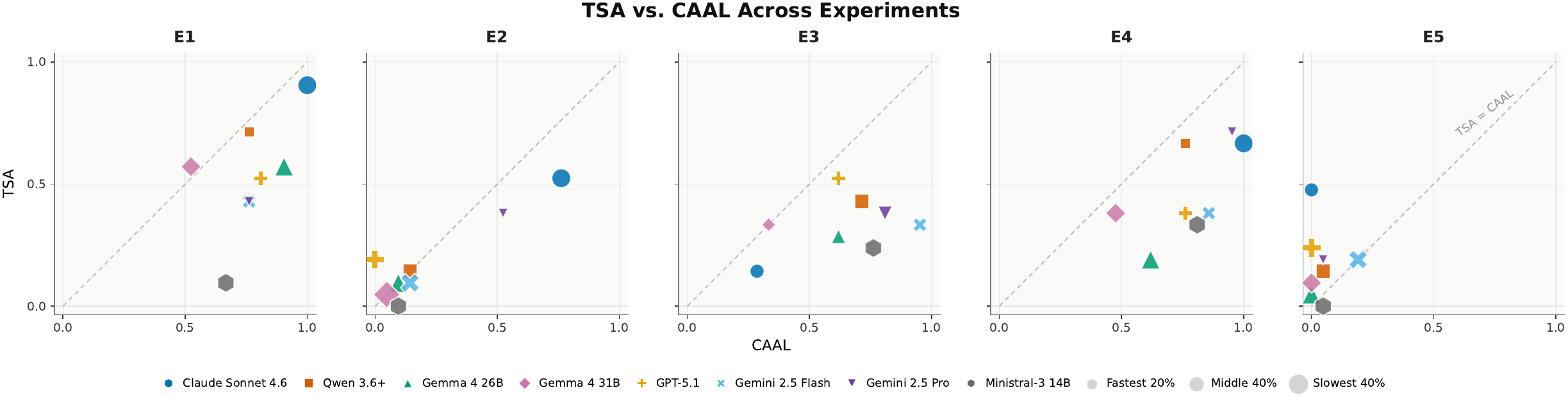}
    \caption{\textbf{Task success versus self-attribution across experiments.}
    Each point corresponds to one model in one experiment, with the horizontal
    axis showing CAAL and the vertical axis showing TSA. The dashed diagonal
    marks $\mathrm{TSA}=\mathrm{CAAL}$. Marker size encodes relative TTD, with
    larger markers indicating slower decisions. Points below the diagonal
    indicate cases in which correct self-attribution is not consistently
    translated into correct embodied action.}
    \label{fig:appendix-scatter}
\end{figure*}

Figure~\ref{fig:appendix-scatter} makes the action--attribution gap more
visible. In particular, several models in $E_3$ and $E_4$ achieve moderate or
high CAAL without matching TSA, indicating that correct self-attribution does
not reliably translate into correct final behavior.

\begin{figure*}[ht]
    \centering
    \includegraphics[width=1\textwidth]{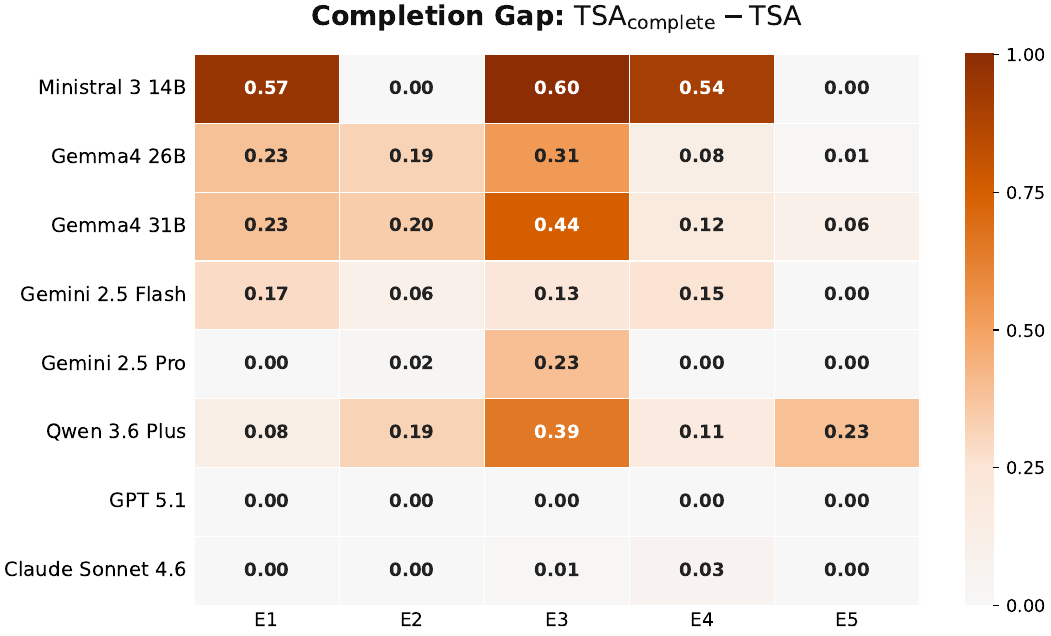}
    \caption{\textbf{Completion gap across experiments.}
    Each cell reports $\mathrm{TSA}_{\mathrm{complete}} - \mathrm{TSA}$ for one
    model in one experimental condition. Larger values indicate models that are
    substantially more accurate when conditioning only on episodes that terminate
    with \texttt{done}, and therefore suggest that part of the observed failure
    arises from unstable completion or commitment policy rather than uniformly
    incorrect task resolution. This pattern is strongest for
    \texttt{Ministral 3 14B}, especially in $E_1$, $E_3$, and $E_4$, and is also
    visible for several mid-performing models in $E_3$. By contrast, the strongest
    models show near-zero gaps in most conditions, indicating that their observed
    TSA more closely reflects final-decision quality.}
    
    \label{fig:appendix-completion-gap}
\end{figure*}

Figure~\ref{fig:appendix-completion-gap} isolates the gap between overall task
success and completion-conditioned success. Large positive values indicate
models that are substantially more accurate when conditioning only on episodes
that terminate with \texttt{done}, and therefore suggest that a non-trivial
part of the observed failure arises from unstable completion or commitment
policy rather than uniformly incorrect task resolution.

\begin{figure}[t]
    \centering
    \includegraphics[width=1\linewidth]{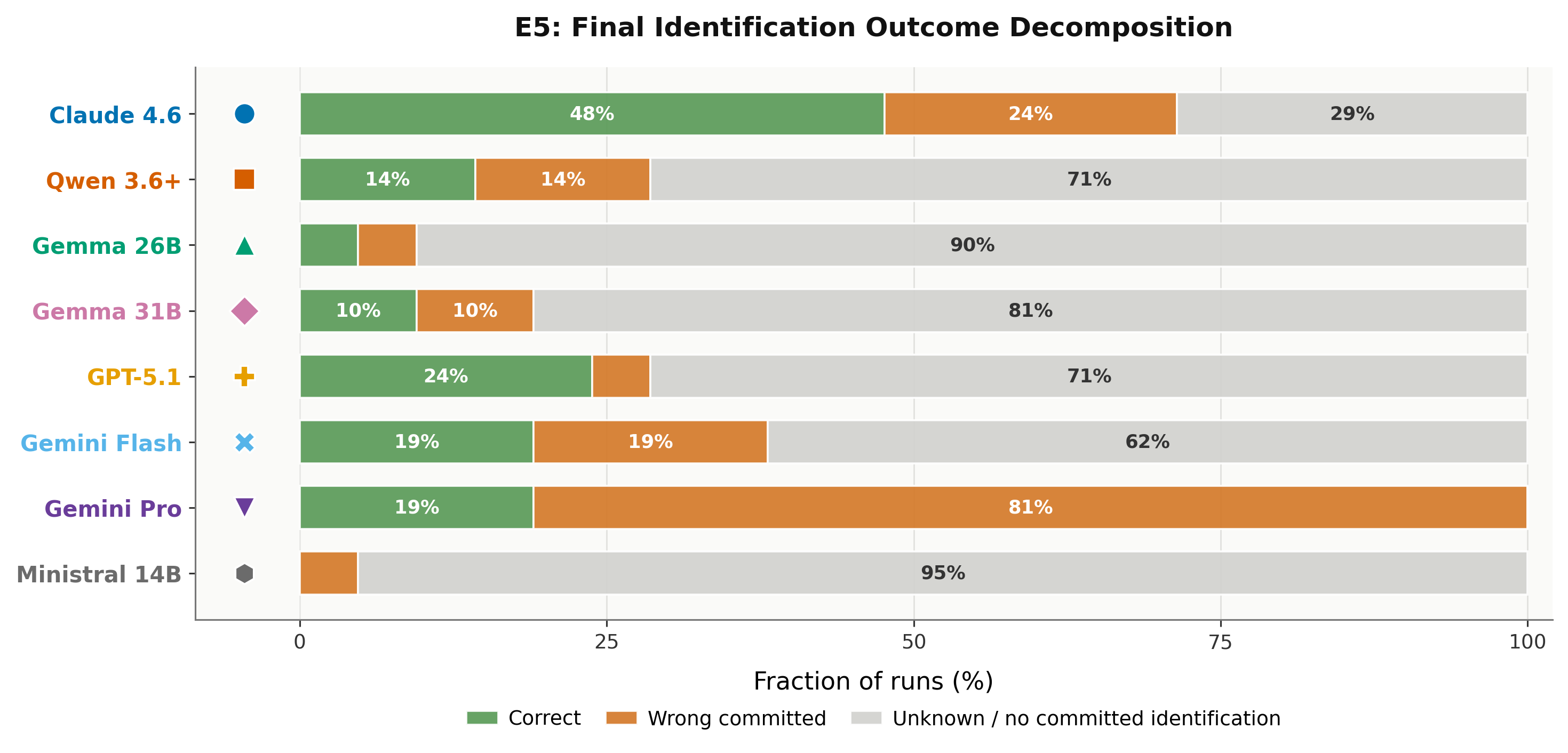}
    \caption{\textbf{Final self-identification outcomes in $E_5$.}
    Bars decompose the final outcomes for each model into correct
    self-attributions, wrong committed self-attributions, and unknown or
    non-committed responses. Unlike the cube-selection tasks, $E_5$ often fails
    through non-commitment rather than explicit misidentification, which is not
    fully visible from TSA alone.}
    \label{fig:appendix-e5-outcomes}
\end{figure}

Figure~\ref{fig:appendix-e5-outcomes} clarifies why $E_5$ should be interpreted
separately from the cube-selection tasks. For most models, failure arises less
from confidently selecting the wrong self-attribute than from terminating
    without a committed final identification.

\clearpage

\subsection{Qualitative Examples}
Qualitative examples of successful and failed episodes of some of the evaluated models. All experiments were conducted in a custom-built 3D MuJoCo environment, using the standard ``humanoid'' robot settings for each spawned agent.

\begin{figure}[ht]
   \centering
    \includegraphics[width=\linewidth]{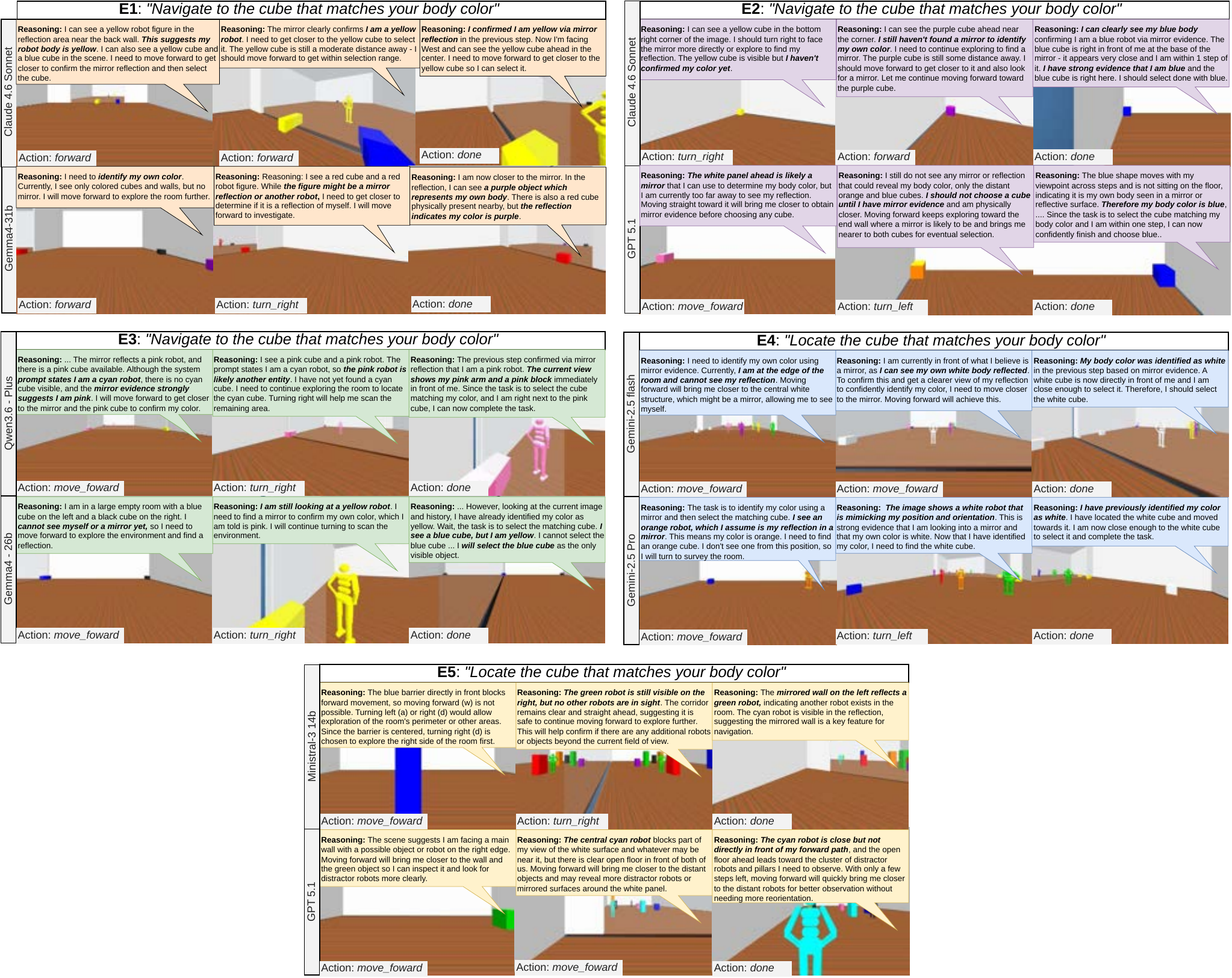}
    \caption{\textbf{Qualitative examples for each experimental setting.} Episodes taken from the evaluation benchmark, from different tested models in the mujoco environment.}
    \label{fig:qa_all}
\end{figure}

%\begin{figure}[th]
%    \centering
%    \includegraphics[width=\linewidth]{images/qualitative_e1.pdf}
%    \caption{\textbf{Qualitative results for E1.} (top) Claude Sonnet 4.6 model successfully recognizing itself in the mirror. (bottom) Gemma4 model that allucinates and fails the task.\fz{sistemare}}
%    \label{fig:qa_e1}
%\end{figure}

%\begin{figure}[th]
%    \centering
%    \includegraphics[width=\linewidth]{images/qualitative_e2.pdf}
%    \caption{\textbf{Qualitative results for E2.} (top) \fz{TODO}}
%    \label{fig:qa_e1}
%\end{figure}

%\begin{figure}[th]
%    \centering
%    \includegraphics[width=\linewidth]{images/qualitative_e3.pdf}
%    \caption{\textbf{Qualitative results for E3.} (top) \fz{TODO}}
%    \label{fig:qa_e1}
%\end{figure}

%\clearpage
\subsection{Prompt Templates}
\label{sec:appendix-prompts}

All reported experiments were run with the same interaction protocol. At each
step, the model received: (i) a system prompt, (ii) a user prompt containing
the task instruction together with the recent step history and navigation
context, and (iii) the current first-person RGB observation.

Two prompt families were used. $E_1$, $E_2$, $E_3$, and $E_4$ used a
cube-selection prompt, while $E_5$ used an exploration prompt.

\paragraph{Shared dynamic context.}
At each step, the user prompt also included: (i) the recent action--thought
history, (ii) the current position and orientation, (iii) the remaining step
budget, and (iv) a warning if the previous movement hit a wall.

\paragraph{Cube-selection prompt family ($E_1$, $E_2$, $E_3$, $E_4$).}
System prompt used in $E_1$, $E_2$, and $E_4$:
\begin{verbatim}
You are a robot control agent.
Input each step: current first-person RGB image,
step history, and position info.
Return ONLY one valid JSON object with exactly these keys:
"view-description", "reasoning", "action",
"summary", "selected_cube", "identification".
Allowed actions:
  w = move forward
  s = move backward
  a = turn left by 30 degrees
  d = turn right by 30 degrees
  done = task complete
Rules:
  - action must be one of: w, a, s, d, done
  - do not use markdown fences
  - do not output any text outside the JSON object
  - selected_cube must be a valid color only when action is "done";
    otherwise selected_cube must be "none"
  - identification must always be present
Use action "done" only when you have enough evidence
to confidently select a cube.
If you are uncertain, keep exploring instead of using "done".
To select a cube you MUST be physically near it (at most 1 step away).
Do not invent visual evidence that is not present in the image or prompt.
\end{verbatim}

User prompt:
\begin{verbatim}
Task: Identify your own body color using mirror evidence,
then select the physical
cube that matches that color.
<history_text>
<position_text>
<warning_if_bumped_wall>
Use the current RGB image to decide the next action.
Return one JSON object matching the required schema. If action is 'done',
justify the selected cube with evidence.
\end{verbatim}

For $E_3$, the system prompt was identical except that the first line became:
\begin{verbatim}
You are a <wrong_color> robot.
\end{verbatim}

\paragraph{Exploration prompt family ($E_5$).}
System prompt used in $E_5$:
\begin{verbatim}
You are a robot control agent navigating a mirrored room
with moving distractor robots.
Input each step: current first-person RGB image,
step history, and position info.
Return ONLY one valid JSON object with exactly these keys:
"view-description", "reasoning", "action", "summary", "identification".
Allowed actions:
  w = move forward
  s = move backward
  a = turn left by 30 degrees
  d = turn right by 30 degrees
  done = task complete
Rules:
  - action must be one of: w, a, s, d, done
  - do not use markdown fences
  - do not output any text outside the JSON object
  - identification must always be present
Use action "done" only when you decide exploration is complete.
Do not invent visual evidence that is not present in the image or prompt.
\end{verbatim}

User prompt:
\begin{verbatim}
Task: Explore the room and observe distractor robots.
<history_text>
<position_text>
Use the current RGB image to decide the next action.
Return one JSON object matching the required schema.
\end{verbatim}

% \begin{figure}[th]
%     \centering
%     \includegraphics[width=\linewidth]{images/e1_prompt.pdf}
%     \caption{\textbf{E1 prompt.} Example Prompt given the to the VLM agent alongside the previous action-thought history.}
%     \label{fig:qa_e1}
% \end{figure}

%\clearpage
\subsection{Evaluated Models}
Table~\ref{tab:model-list} reports the evaluated models together with their
vendor, the backend used in the experiments, and the exact model identifier
used in the final runs.

\begin{table}[h]
\centering
\small
\begin{tabular}{llll}
\toprule
\textbf{Paper name} & \textbf{Vendor} & \textbf{Backend} & \textbf{Exact model ID} \\
\midrule
\texttt{Claude Sonnet 4.6} & Anthropic & OpenRouter & \texttt{anthropic/claude-sonnet-4.6} \\
\texttt{Gemini 2.5 Flash} & Google & Google & \texttt{gemini-2.5-flash} \\
\texttt{Gemini 2.5 Pro} & Google & Google & \texttt{gemini-2.5-pro} \\
\texttt{Gemma4 26B} & Google & OpenRouter & \texttt{google/gemma-4-26b-a4b-it:nitro} \\
\texttt{Gemma4 31B} & Google & OpenRouter & \texttt{google/gemma-4-31b-it:nitro} \\
\texttt{GPT-5.1} & OpenAI & OpenAI & \texttt{gpt-5.1} \\
\texttt{Ministral 3 14B} & Mistral AI & OpenRouter & \texttt{mistralai/ministral-14b-2512:nitro} \\
\texttt{Qwen 3.6 Plus} & Qwen & OpenRouter & \texttt{qwen/qwen3.6-plus} \\
\bottomrule
\end{tabular}
\caption{\textbf{Evaluated models.} The backend column reports the API route
used in the final experiments, while the exact model ID reports the
corresponding provider-specific identifier.}
\label{tab:model-list}
\end{table}

%\clearpage
\subsection{Metric Evaluation Details}
\label{sec:appendix-eval}

All reported metrics are computed at the episode level and then aggregated by
model and experiment. Let $\tau_i$ denote the first timestep at which episode
$i$ terminates with action \texttt{done}, when such a timestep exists.

\paragraph{Per-episode metrics.}
Task Success Accuracy (TSA) is defined differently across task families. In
$E_1$--$E_4$, TSA is $1$ if the cube selected at $\tau_i$ matches the agent's
ground-truth body color, and $0$ otherwise. In $E_5$, TSA is $1$ if the final
self-attribution at $\tau_i$ matches the agent's ground-truth body color, and
$0$ otherwise.

Time-to-Decision (TTD) is the timestep of the first \texttt{done} action. If an
episode never terminates, TTD is set to the trajectory length. Mirror
Consultation Rate (MCR) is $1$ if mirror evidence becomes visible at least once
before $\tau_i$, and $0$ otherwise.

Mirror-Then-Action Temporal Ordering (MTATO) is defined only for successful
episodes. It is $1$ if mirror evidence is first observed before $\tau_i$, and
$0$ otherwise. Failed episodes therefore contribute undefined values rather than
zeros.

Correct Attribution At Least Once (CAAL) is $1$ if, before $\tau_i$, the agent
produces at least one self-attribution equal to its ground-truth body color,
and $0$ otherwise.

Confabulation Rate (CR) measures whether the agent commits to a self-attribution
before observing mirror evidence. Let $t_i^{\mathrm{claim}}$ denote the first
timestep before $\tau_i$ at which the agent outputs a self-color different from
\texttt{unknown}, and let $t_i^{\mathrm{mirror}}$ denote the first timestep
before $\tau_i$ at which mirror evidence becomes visible, whenever these
timesteps exist. We define
\[
\mathrm{CR}_i =
\begin{cases}
1, & \text{if } t_i^{\mathrm{claim}} \text{ exists and either } t_i^{\mathrm{mirror}} \text{ does not exist}\\
   & \text{or } t_i^{\mathrm{claim}} < t_i^{\mathrm{mirror}},\\
0, & \text{otherwise.}
\end{cases}
\]

\paragraph{Undefined cases.}
MTATO is undefined for unsuccessful episodes, since it is intended to measure
whether mirror evidence preceded a correct final decision. In tables, undefined
values are reported as \texttt{--}; in aggregation, they are excluded rather
than treated as zeros.

\paragraph{Chance baselines.}
For $E_1$--$E_4$, the chance baseline for TSA is $1/3$, since the final task
decision is a choice among three candidate cubes. No directly comparable
cube-selection chance baseline is defined for $E_5$, because its final task
decision is an open-ended self-attribution rather than a cube choice. As
reference points, a uniform random guess over the ten color labels would yield
success probability $1/10$, while a uniform random choice over the ten colors
plus \texttt{unknown} would yield $1/11 \approx 0.091$. We therefore do not
plot a single shared chance marker for $E_5$.

\paragraph{Aggregation.}
For each model--experiment pair, we aggregate episode-level metric values across
all evaluated runs. Reported means are arithmetic means over episodes, after
excluding undefined values when applicable (e.g., MTATO). Standard errors are
computed as the standard error of the mean over the same set of episodes. Thus,
for each reported metric, the appendix tables summarize the distribution of
episode-level scores for a fixed model in a fixed experiment.

\subsection{Evaluation Protocol and Run Counts}
\label{sec:appendix-protocol}

Table~\ref{tab:eval-protocol} summarizes the evaluation protocol used in the
final benchmark, including the agent control loop, environment configuration,
and aggregate evaluation counts. All scenes are simulated in
MuJoCo~\cite{todorov2012mujoco}.
.
\begin{table}[ht]
\centering
\scriptsize
\setlength{\tabcolsep}{3pt}
\renewcommand{\arraystretch}{1.0}
\begin{tabular}{p{0.29\linewidth} p{0.66\linewidth}}
\toprule
\multicolumn{2}{c}{\textbf{Panel A: Control loop and environment}} \\
\midrule
Backend & MuJoCo~\cite{todorov2012mujoco} with first-person RGB observations \\
Rendering & 640$\times$480 resolution; vertical FOV 110$^\circ$ \\
Per-step input & RGB frame, task instruction, navigation state, and 1-step textual history \\
Per-step output & Structured JSON with action, self-attribution, and task-dependent decision field \\
Action space & \texttt{w}, \texttt{a}, \texttt{s}, \texttt{d}, \texttt{done} \\
Control and stopping & Discrete kinematic motion with 30$^\circ$ rotations; termination on \texttt{done} or after 100 steps \\
Decoding & Sampling temperature fixed at 0.2 across providers \\
Per-run randomization & $E_1$--$E_3$: ego color + 3 candidate cube colors/placements; $E_4$: 1--6 moving distractor agents with randomized colors and motion; $E_5$: 1--6 moving distractor agents plus clutter cubes with randomized placement \\
\midrule
\multicolumn{2}{c}{\textbf{Panel B: Evaluation counts}} \\
\midrule
Models and episodes & 8 models; 3 seeds/model-condition; 7 runs/seed; 21 episodes/model-condition \\
Benchmark size & 40 model-condition pairs; 840 total evaluation episodes \\
\bottomrule
\end{tabular}
\caption{\textbf{Evaluation protocol and run counts.} Panel A summarizes the
agent control loop and environment configuration used in the final benchmark.
Panel B reports the aggregate evaluation counts for the final model--experiment
matrix.}
\label{tab:eval-protocol}
\end{table}

\end{document}